%% file: wang2017_arxiv.tex
\documentclass[runningheads]{llncs}
\usepackage{graphicx}
\usepackage{amsmath,amssymb}
\usepackage{color}
\usepackage[width=122mm,left=12mm,paperwidth=146mm,height=193mm,top=12mm,paperheight=217mm]{geometry}

\usepackage{xfrac}
\newcommand{\citep}{\cite}
\newcommand{\citet}{\cite}
\usepackage{pgfplots}
\pgfplotsset{compat=newest}

\usepackage[]{algorithm2e}

\graphicspath{}
\newcommand{\folder}{.}
\newdimen{\figurewidth} 
\begin{document}
\pagestyle{headings}
\mainmatter

\title{Viewpoint Adaptation for Rigid Object Detection}

\titlerunning{Viewpoint Adaptation for Rigid Object Detection}

\authorrunning{Wang et al.}

\author{Patrick Wang\inst{1,2}, Kenneth Morton\inst{2}, Peter Torrione\inst{2}, and Leslie Collins\inst{1}}
\institute{Duke University
\and CoVar Applied Technologies\\
\email{\{patrick.wang, leslie.collins\}@duke.edu\\ \{kenny, pete\}@covar.com}}

\maketitle

\begin{abstract}
An object detector performs suboptimally when applied to image data taken from a viewpoint different from the one with which it was trained. In this paper, we present a viewpoint adaptation algorithm that allows a trained single-view object detector to be adapted to a new, distinct viewpoint. We first illustrate how a feature space transformation can be inferred from a known homography between the source and target viewpoints. Second, we show that a variety of trained classifiers can be modified to behave as if that transformation were applied to each testing instance. The proposed algorithm is evaluated on a person detection task using images from the PETS 2007 and CAVIAR datasets, as well as from a new synthetic multi-view person detection dataset. It yields substantial performance improvements when adapting single-view person detectors to new viewpoints, and simultaneously reduces computational complexity. This work has the potential to improve detection performance for cameras viewing objects from arbitrary viewpoints, while simplifying data collection and feature extraction.

\keywords{viewpoint, domain, adaptation, perspective, projection, object, detection}
\end{abstract}

\section{Introduction}
The automated detection of objects in images is an important task in a wide variety of applications: autonomous driving, robotics, surveillance, image annotation, and others. While the problem to be solved in each of these scenarios is fundamentally the same, each scenario has different implicit expectations for the position and orientation of the camera with respect to objects in the scene.

\begin{figure}[tb]
\begin{center}
   \includegraphics[width=0.5\linewidth]{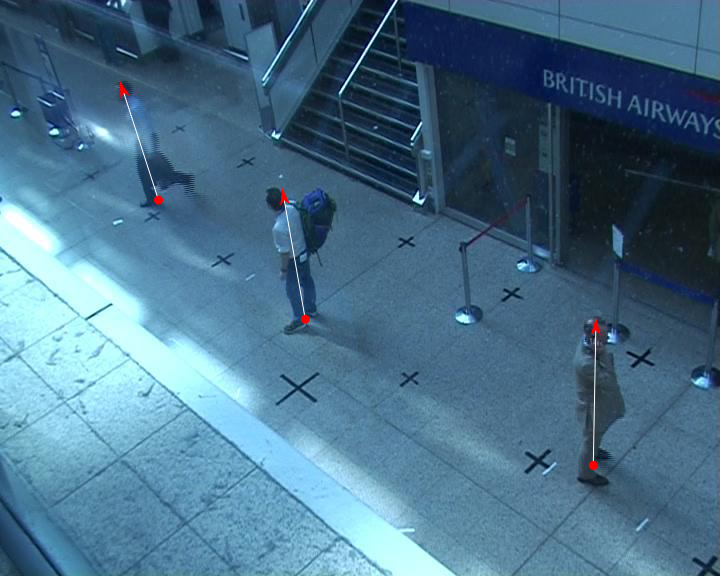}
\end{center}
   \caption{\label{fig:perspectivePersons}
Example surveillance camera image from the PETS 2007 dataset (camera 3) \citep{pets2007data}. The size, orientation, and proportions of person images depend on their position and orientation relative to the camera. Line segments connecting the persons' heads and feet illustrate variations in orientation.}
\end{figure}

Recent work, for example, has developed and refined algorithms for pedestrian detection, i.e. the detection of persons from a camera mounted on the hood of a car \citep{benenson2013seeking,dollar2014fast}. With this common camera pose, these detectors typically view persons only one way: horizontally at about waist level. This viewpoint is reflected in the popular datasets used to evaluate pedestrian detection algorithms \citep{dalal2005histograms,dollar2012pedestrian}. Pedestrian detectors will not perform optimally if applied to, for example, a mall security camera, which may see persons from a wide variety of relative viewpoints (see Figure~\ref{fig:perspectivePersons}). One goal of this work is to leverage camera pose information along with the existing body of work in pedestrian detection to develop fast and accurate person detectors for cameras with arbitrary (known) poses.

More generally, it is desirable to be able to adapt any rigid object detector trained for one viewpoint to another distinct viewpoint. This is an instance of the ``domain adaptation'' problem, in which the conditions under which a detector is to be applied do not match the those under which the training data were collected \citep{gopalan2011domain}. ``Domain'' can generally refer to a variety of factors like viewpoint, lighting, and resolution. 

Unlike other common domain factors, however, viewpoint effects cannot easily be normalized through feature extraction. We will show that, instead, information about the object and camera geometry can be used to decouple the effects of viewpoint from other domain factors and to learn a feature space transformation between the source and target viewpoints. The feature transformation can then be used to adapt a trained object detector to the target viewpoint. This novel adaptation procedure enables improved detection across different viewpoints while requiring very little additional run-time computation. Furthermore, our approach does not require training data from the target viewpoint, in contrast with existing domain adaptation algorithms.

\section{Related Work}
Many domain adaptation algorithms explicitly learn a feature space transformation using data from the source and target viewpoints \citep{kulis2011you}. Others attempt to select or re-weight training examples from one or more source viewpoints so that they approximately reflect the distribution of instances from the target viewpoint \citep{cao2013transfer,dai2007boosting}. These existing domain adaptation algorithms universally require training images from both the source and target viewpoints and they often also require the corresponding labels from one or both domains. For many applications, however, it is impractical to manually collect new data and retrain the detector for every viewpoint that might be encountered during operation. For moving cameras or those with wide-angle lenses capturing a continuous range of relative perspectives, traditional domain adaptation will be impossible.

Some object detectors instead extract features designed to be invariant to small geometric transformations of various types. For example, scale-invariant feature transforms (SIFT) \citep{lowe1999object} are partially invariant to changes in objects' scale. Histograms of oriented gradients (HOG) \citep{dalal2005histograms} are invariant to small local rotations. Such viewpoint-invariant local features are important to help normalize pose differences in deformable objects. They cannot, however, capture large-scale shape changes, like those illustrated in Figure~\ref{fig:perspectivePersons}.

For situations like this, many existing methods rely on machine learning algorithms to handle viewpoint differences. These require training data from all viewpoints of interest, and use robust classifiers to distinguish between the (very diverse) class of target instances and the class of background instances. The classification problem resulting from such ``multi-view'' approaches may be much more complex than detecting persons from only a single viewpoint. Furthermore, it is usually difficult to collect training data sets containing a comprehensive range of perspectives.

Some approaches attempt to avoid these issues by using a 3D model along with images from a small number of ``canonical'' views to detect objects with arbitrary relative poses \citep{danielsson2011projectable,savarese2008view,thomas2006towards}. These do not take advantage of any known camera pose information, and can suffer from high runtime computational requirements.

For ease of computation, the assumption of a planar model is common in optical character recognition \citep{howe2005boosted,miyao2006virtual} and has also been used in face recognition \citep{tan2006face} and in some general object detection approaches \citep{kalal2012tracking}. Li et al. \citet{li2008human} apply this planar assumption to person detection, allowing them to warp candidate detection windows to match the trained classifier's viewpoint (Figure~\ref{fig:camera}). They show that significant performance improvements may be obtained with this method compared to using a single-view classifier without compensatory warping. However, image warping involves (at run-time) extracting features from each candidate detection window independently, which is impractical for real-time applications.

In contrast, we approach the problem by modifying the trained detector itself to fit the viewpoint in which it will be used. The only other work in computer vision that transforms detection models focuses exclusively on scale \citep{dollar2014fast,benenson2012pedestrian}.

\section{Viewpoint Adaptation}
Perspective mapping approaches can be used for object detection scenarios in which information is available about the relative geometry of camera and object. This is the case for most person detection applications because camera geometry is typically parameterized with respect to some point on the ground plane, and persons can usually be found adjacent and normal to the ground plane. Many other objects follow similar geometric rules, includings cars, furniture, and many animals.

In contrast with the typical approach in which a projective warping is applied to each testing window before computing features (as in \citet{li2008human}, Algorithm~\ref{alg:li}), we approximate this process by instead applying an appropriate linear transformation to the features themselves (Algorithm~\ref{alg:adapt}). For some classes of detectors, we can further improve computational efficiency by observing that this feature transformation is equivalent to a suitable transformation of the trained classifier parameters.

The classifier can thus be adapted to the target viewpoint, requiring nothing but a trained single-view classifier and a homography relating the source and target viewpoints through the presumed object plane. This approach has the benefit of needing only training data from a single viewpoint, so that extensive effort need not be spent acquiring diverse training images. In particular, no data is required from the specific viewpoint in which the classifier will be applied; it is sufficient to have knowledge of the target camera's pose.

\subsection{Camera Homography}
\begin{figure}[tb]
\begin{center}
  \includegraphics[width=0.45\linewidth]{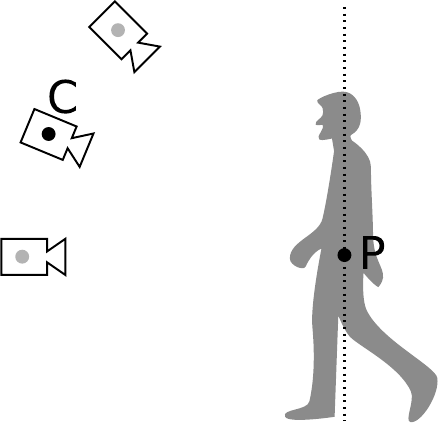}
  \quad
	\input{\folder/20160304_imageWarpingExample.tex}
\end{center}
   \caption{\label{fig:camera}
Left, the person is assumed to be planar, lying on the plane indicated by the dotted line. Three viewpoints are considered, from cameras at elevations of $0$, $\sfrac{\pi}{8}$, and $\sfrac{\pi}{4}$ radians. Right, images from each elevation are warped to approximate $0$ elevation (no transformation is necessary for the $0\rightarrow 0$ case). The approximation degrades as elevation increases due to inaccuracies in the planar human model. }
\end{figure}

The object is assumed to be planar and perpendicular to the projection of the line segment $CP$ onto the ground plane, where $C$ is the camera's position and $P$ is the object's position. This plane is illustrated by the dotted line in Figure~\ref{fig:camera}. A point on this object plane is described using homogeneous coordinates by $\boldsymbol{u}_p = [u_p,v_p,1]^T$. The corresponding point in image $i$, $\boldsymbol{u}_i = [u_i,v_i,1]^T$, is related to the object-plane point by $\boldsymbol{u}_i = H_i\boldsymbol{u}_p$, where $H_i$ is a 3-by-3 homography matrix. The relationship between points in images from two cameras with different viewpoints is then given by $\boldsymbol{u}_1 = H_1H_2^{-1}\boldsymbol{u}_2 = H_{12}\boldsymbol{u}_2$.

\setlength{\figurewidth}{0.33\linewidth}

\subsection{Local Feature Re-mapping}
Many object detection algorithms can be defined in terms of a set of local image feature descriptors that are computed for each frame, and a sliding window classifier that uses a small set of those features to make a detection decision \citep{dollar2014fast,dalal2005histograms,viola2001rapid,krizhevsky2012imagenet}. Feature re-mapping can be applied with most detectors of this type, given the homography ($H_{12}$, previously) between source- and target-domain windows. We will demonstrate this viewpoint adaptation for two general classes of local features, which we will call Haar-like and HOG-like. Haar-like and HOG-like features cover a wide variety of the feature types used in modern object detectors \citep{dollar2014fast,dalal2005histograms,viola2001rapid,felzenszwalb2010object}.

\begin{figure}[tb]
\begin{center}
   \includegraphics[width=0.45\linewidth]{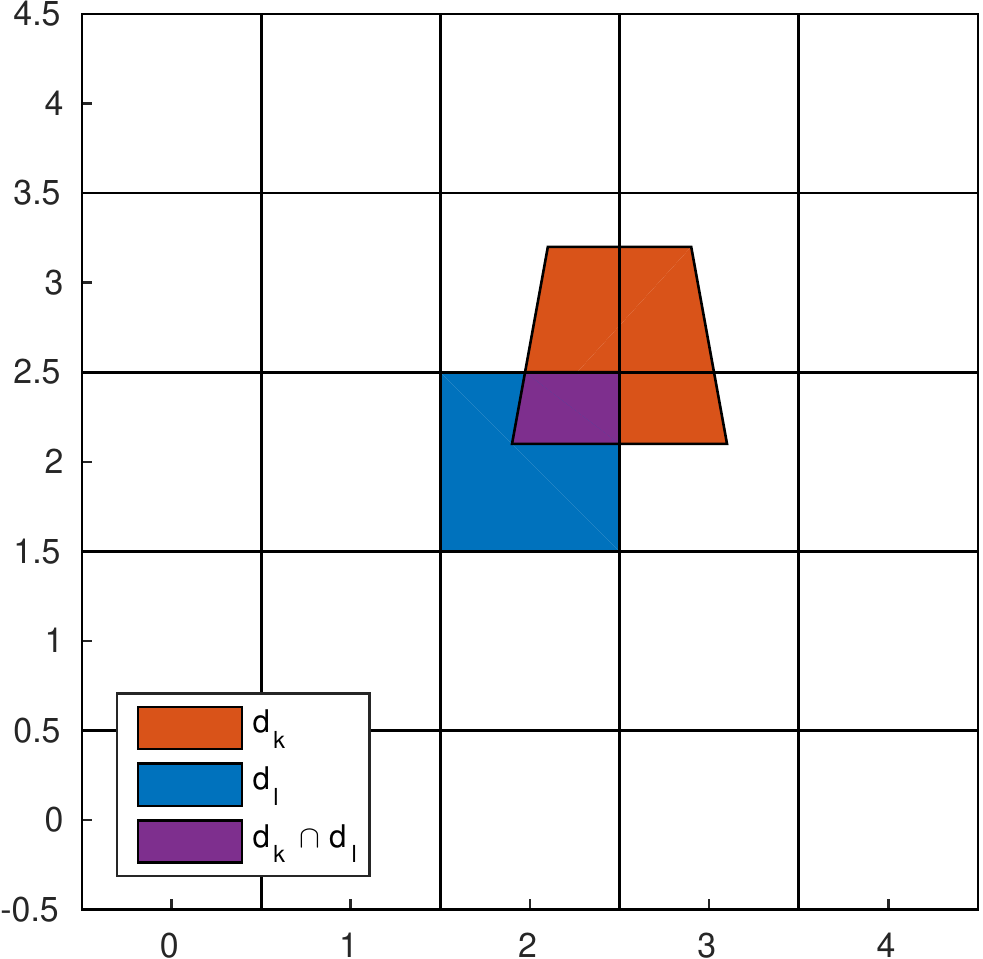}
   \includegraphics[width=0.45\linewidth]{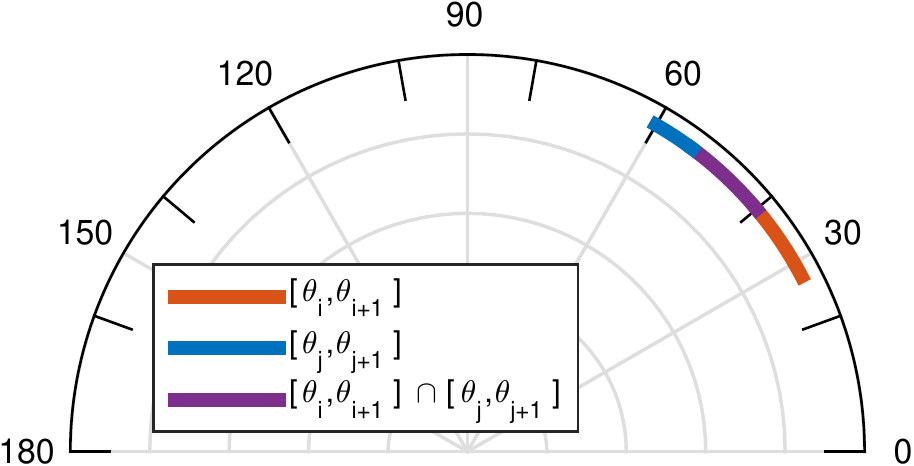}
\end{center}
   \caption{\label{fig:cellResampling}
Left, each cell $d_l$ from the first viewpoint is projected to the corresponding location $d_k$ in the second viewpoint. Right, each angle histogram bin $[\theta_i,\theta_{i+1}]$ from the first viewpoint is projected to the corresponding location $[\theta_j,\theta_{j+1}]$ in the second viewpoint. The relative overlap of these elements is used to generate a linear feature space mapping.}
\end{figure}

The term ``Haar-like feature'' is used here to describe any local image feature computed as the sum of pixel intensities over a rectangular image region (a ``cell''). It also applies to sums of any pixel-wise function of the image intensity (e.g. sums of squared pixel intensities). These cells are frequently computed in a grid covering the image, as in Figure~\ref{fig:cellResampling} (left). Each cell has an ``extent'' $d_k=\begin{bmatrix}\boldsymbol{u}_{\textrm{NE}}^T& \boldsymbol{u}_{\textrm{SE}}^T& \boldsymbol{u}_{\textrm{SW}}^T& \boldsymbol{u}_{\textrm{NW}}^T\end{bmatrix}$ defined by the coordinates of its corners. To compute the Haar-like feature from a cell in the target viewpoint given the Haar-like features computed from the source (trained) viewpoint,  the target-cell extent is transformed via $d_{k'} = H_{st}d_k$ using the appropriate source-target homography $H_{st}$. This extent is then overlaid on the computed (source-view) cell grid, and the result is the sum of the values of the overlapped cells, weighted by the degree to which they overlap the projected extent. Thus the contribution from source-view cell $l$ to target-view cell $k$ is given by
\begin{equation}
    S_{kl} = \frac{\text{area}\left(d_l\cap d_{k'}\right)}{\text{area}\left(d_l\right)}
\end{equation}
This is illustrated in Figure~\ref{fig:cellResampling} (left).

\begin{figure}[tb]
\begin{center}
   \includegraphics[width=0.9\linewidth]{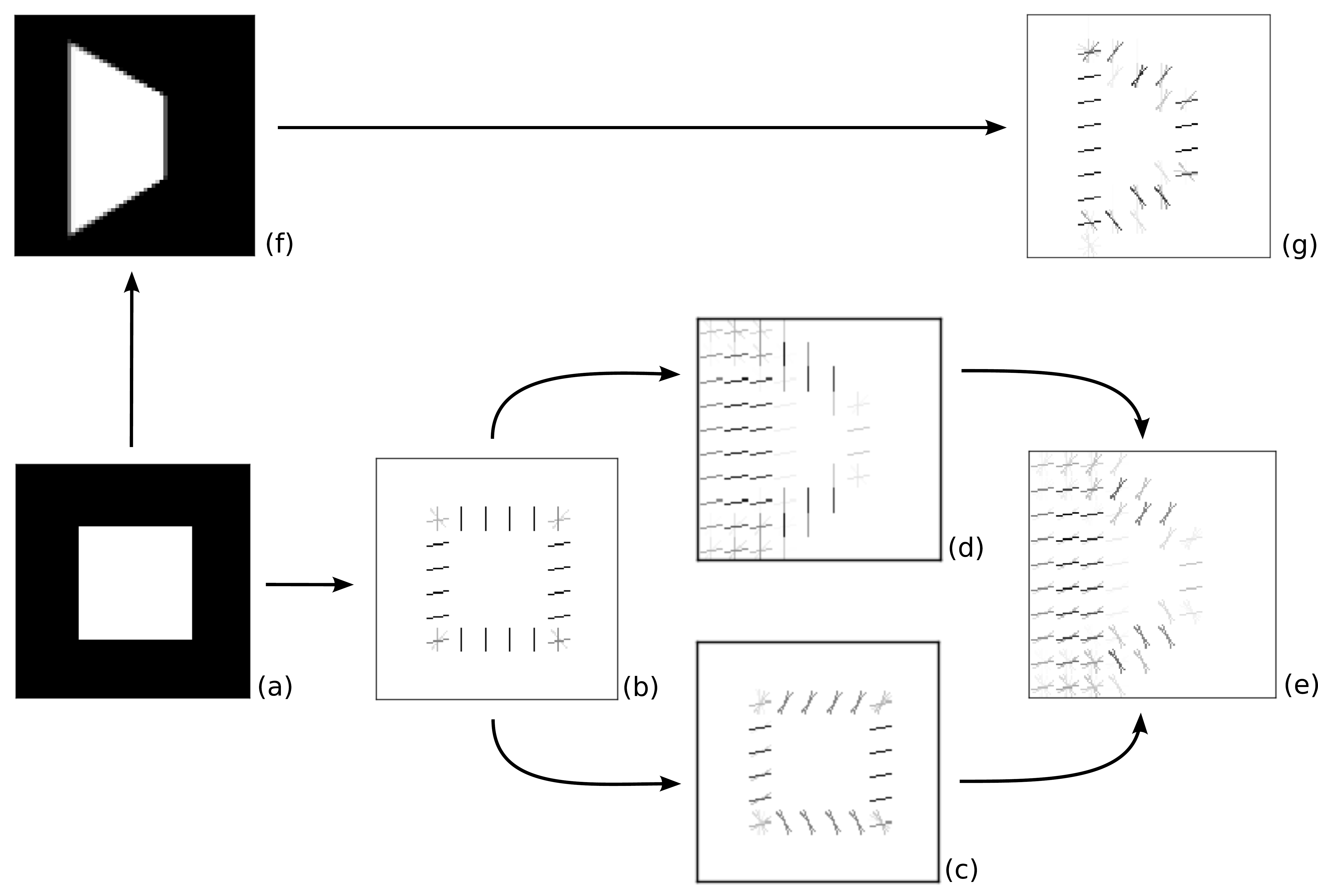}
\end{center}
   \caption{\label{fig:squareExample}
(a) Example image; (b) HOG features from image; (c) HOG features with histograms resampled; (d) HOG features with cells resampled; (e) HOG features with both histograms and cells resampled; (f) image warped; (g) HOG features from warped image. (e) provides an approximation to (g) without requiring the construction of (f).}
\end{figure}

HOG-like features present a slightly greater challenge. They are constructed by computing the gradient (magnitude and angle) across the image, and computing a magnitude-weighted angle histogram within each cell. HOG features for a simple image are shown in Figure~\ref{fig:squareExample}b. HOG gradient angle bins are restricted to a $180^\circ$ arc, so that parallel edges with opposite directions are considered to have the same gradient angle and all gradient magnitudes are positive.

HOG feature re-mapping proceeds in two steps: 1) re-mapping the cell locations like for Haar-like features, and 2) re-mapping the angle histogram bins. The second step requires that we infer an angle transformation from the homography $H_{st}$, and perform a linear resampling over the angle histogram bins similar to what we did for Haar-like features.

Given a homography matrix of the form
\begin{equation}
    H_{st} = \left[ \begin{array}{ccc}
    h_1 & h_2 & h_3\\
    h_4 & h_5 & h_6\\
    h_7 & h_8 & 1
\end{array} \right],
\end{equation}
the gradient angle at a point $[x',y']$ in the target viewpoint is given by
\begin{equation}
	\theta' = \tan^{-1}\left[\frac{(h_4 - h_7 y')\cos(\theta) + (h_5 - h_8 y')\sin(\theta)}{(h_1 - h_7 x')\cos(\theta) + (h_2 - h_8 x')\sin(\theta)}\right]
\end{equation}
where $\theta$ is the corresponding gradient angle in the source viewpoint. This expression is used to map each angle histogram bin edge into the target domain. Then the contribution of computed bin $j$ to desired bin $i$ is given by
\begin{equation}
    A_{ij} = \frac{\text{area}\left([\theta_j,\theta_{j+1}]\cap [\theta_{i'},\theta_{i'+1}]\right)}{\text{area}\left([\theta_j,\theta_{j+1}]\right)}
\end{equation}
This is illustrated in Figure~\ref{fig:cellResampling} (right).


Combining cell resampling and histogram resampling, a HOG feature vector $\boldsymbol{x}'$ can be warped into a new viewpoint domain according to
\begin{equation}
	\boldsymbol{x}'\approx (S\otimes A)\boldsymbol{x} = G\boldsymbol{x},
\end{equation}
where $S$ performs the spatial resampling of the HOG cells and $A$ performs the resampling of the angle histogram bins. $\otimes$ denotes the Kronecker product. The two-step HOG feature re-mapping process is illustrated in Figure \ref{fig:squareExample}a-\ref{fig:squareExample}e. We can see that the proposed feature-remapping algorithm approximates the result of applying an inverse perspective mapping to the image before feature computation (Figure~\ref{fig:squareExample}g).

\begin{algorithm}[tb]
 Given an image $I$\;
 \For{window $w$ in $I$}{
  apply image warping $w \rightarrow w'$\;
  compute features $f(w')$\;
  compute decision statistic\;
 }
 \caption{Viewpoint adaptation by image warping \citep{li2008human}}
 \label{alg:li}
\end{algorithm}

\begin{algorithm}[tb]
 Given an image $I$\;
 compute features $f(I)$\;
 \For{window $w$ in $I$}{
  apply feature re-mapping $f(w)\rightarrow f(w)' \approx f(w')$\;
  compute decision statistic\;
 }
 \caption{Viewpoint adaptation by feature re-mapping}
 \label{alg:adapt}
\end{algorithm}

\section{Classifier Adaptation}
\label{sec:classifierWarping}
The generic approach to applying the proposed viewpoint adaptation with a trained classifier is to independently re-map each of the computed features for each candidate detection window prior to applying the classifier (see Algorithm~\ref{alg:adapt}).

Linear classifiers are an interesting special case for geometric viewpoint adaptation, because they allow the feature re-mapping to be rolled into the classifier weights themselves, removing even the small run-time computational burden incurred while using nonlinear classifiers.

The learned classifier weights $\boldsymbol{w}$ and the class label $\boldsymbol{y}$ are defined such that
\begin{align}
\boldsymbol{y} &= \boldsymbol{w}^T\boldsymbol{x}\\
&= \boldsymbol{w}^T(G\boldsymbol{x}')\\
&= (G^T\boldsymbol{w})^T\boldsymbol{x}'.
\end{align}
Thus we see that identical results can be obtained by either transforming the testing data to match the source viewpoint or by transforming the classifier weights to match the target viewpoint. This novel result allows the application of a feature space transformation off-line, incurring no run-time computational burden. This viewpoint adaptation method for both linear and nonlinear classifiers with Haar-like and HOG-like features has been evaluated for person detection under varying viewpoints using both real and synthetic imagery.

\section{Experiments}
\label{sec:experiments}
We focus here on two specific detectors whose modes of operation represent distinct, important classes of detectors: the classic HOG/SVM pedestrian detector of \citet{viola2001rapid} is a linear classifier using HOG-like features; and the aggregate channel features detector of \citet{dollar2014fast} is a nonlinear classifier using both Haar-like and HOG-like features.

Both detectors were applied to real data from two established video surveillance datasets. Most existing person image datasets are not suitable for this task, lacking viewpoint diversity and/or camera information. The CAVIAR \citep{caviar2004data} and PETS 2007 \citep{pets2007data} datasets on which our methods were evaluated, however, include camera information and contain a modest range of relative viewpoints. What they lack, as we will see later, is diversity of persons and backgrounds, for many of the relative viewpoints.

Despite their shortcomings, real-data experiments provide evidence for the efficacy of the proposed methods, and elucidate some of the problems and benefits encountered when using them. Among the latter is the ability to define candidate detection windows in terms of the 3D world space rather than the 2D image space.

\subsection{Detection window selection}\label{sec:windows}
Human detection by searching 3D space was first discussed in \cite{li2008human}, and the same principles apply in this work. Building a set of candidate world positions amounts to defining a minimum distance between detections (the detector ``stride'') and applying the camera projection. The image-space detector stride, measured in pixels, is replaced by a real-world stride, measured in meters. This is illustrated well in Figure 3 of \citet{li2008human}. We can compute the desired real-world stride by analogy with the pixel-space stride. Specifically, ACF assumes that for any given image scale, persons are $96$ pixels tall. If persons average $1.75$ meters in the real world, then we assume $\sfrac{1.75}{96}\approx0.018$ meters per pixel. For a desired pixel stride of $8$, we then have a real-world stride of $8\times\sfrac{1.75}{96}=0.146$ meters. Using a fixed real-world stride would yield an infinite number of candidate detection windows in scenes containing the horizon, so we further restrict the detection windows to be no closer than 1 pixel apart in the image space.

It should be noted that each position in the image corresponds to a unique relative viewpoint and associated feature mapping. Thus the detection system must, in general, apply different detection parameters for each image position. These can be computed off-line, however, and simply looked up at run-time.

\subsection{Real data results}
\label{sec:realData}
\begin{figure}[tbp]
\begin{center}
	\input{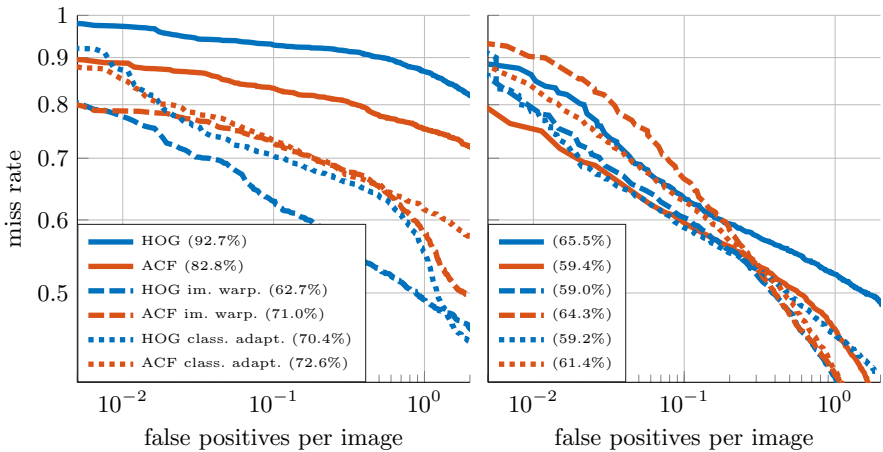}
	\input{\folder/20160401_adaptationRocsPets.tex}
\end{center}
   \caption{\label{fig:realResults}
   Performance on images from the CAVIAR dataset (left) and PETS 2007 dataset (right) is shown for the HOG/SVM detector (blue) and the ACF detector (red). The baseline detectors are shown as solid curves, the image-warping method of \citet{li2008human} is shown as dashed curves, and our classifier adaptation methods are shown as dotted curves. The legend indicates the log-average miss rate between fppi of $10^{-2}$ and $10^0$ for each detector.}
\end{figure}

The PETS 2007 dataset \citep{pets2007data} (camera 3) contains camera calibration information, but ground truth was estimated manually using the \verb|vbb| labeling software \citep{dollar2012pedestrian}. The CAVIAR dataset \citep{caviar2004data} (\verb|Meet_Split_3rdGuy|) comes with ground truth, but the camera was calibrated ahead-of-time using the provided pixel correspondences. The results presented in Figure~\ref{fig:realResults} compare our viewpoint adaptation method for HOG/SVM and ACF to the original, unadapted detectors and to the corresponding image-warping methods \citep{li2008human}. All detectors were trained with the INRIA training data \citep{dalal2005histograms}.

\setlength{\figurewidth}{0.45\linewidth}

For the CAVIAR dataset \citep{caviar2004data}, which has some extreme perspective warping effects, all perspective compensation methods improve substantially over the n{\"a}ive detectors. The sharp decrease in miss rate for some detectors around 1 false positive per image can be attributed to a shelf unit visible in the scene. This specific false alarm can also be seen to cause problems in Figure 9 of \citet{li2008human}. The large impact of a single persistent false alarm highlights an issue with using stationary cameras for evaluation of viewpoint adaptation algorithms: we are essentially given a single source of potential false alarms, whose impact is multiplied by the number of frames in the video. Unfortunately, this issue is unavoidable using existing person detection datasets.

For the PETS 2007 dataset \citep{pets2007data}, which has much more mild perspective effects, both viewpoint adaptation methods yield smaller improvements, even weakening performance for some false positive rates compared to the unadapted detector. This behavior is expected for scenes with insignificant viewpoint changes and is reasonable in this case, given the strong sources of noise described in the previous paragraph.

For both datasets, the adapted ACF detector performs worse than the adapted HOG detector, contrary to our expectations. This unintuitive result requires some explanation. The standard ACF detector achieves excellent detection performance on most pedestrian detection datasets by training a relatively precise detector and applying it at a large number of positions and scales. Contrast this with the HOG detector, which typically uses a larger cell size to achieve a detector more robust to small changes in the position and scale of detection windows. When these two detectors are applied to the same set of manually extracted detection windows in the image warping scenario, HOG is more robust to sources of error including mis-estimation of the camera pose and persons of different-than-average height. These sources of error outweigh the benefits from inverse perspective warping in the PETS 2007 dataset, so we see image warping potentially hurt ACF performance. In the CAVIAR dataset, on the other hand, the perspective effects are severe enough that image warping significantly improves the performance of both detectors.

These real-data experiments begin to demonstrate the capabilities of the proposed methods, but due to the aforementioned dataset issues, are not entirely well-suited to rigorously evaluating viewpoint adaptation methods. To provide an additional, more thorough, evaluation, the HOG/SVM detector was applied to a new synthetic multi-view person detection dataset generated using 3D modelling software.

\subsection{Multi-view dataset}
\begin{figure}[tb]
\begin{center}
   \includegraphics[width=0.5\linewidth]{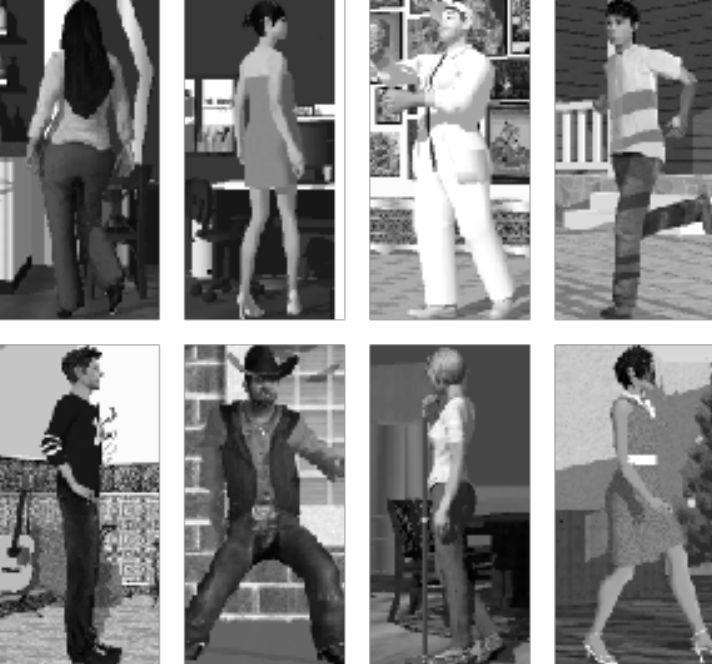}
\end{center}
   \caption{\label{fig:sketchupPersons}
Some example person images generated using SketchUp. The person models are placed randomly in the scene with random rotation about their vertical axis.}
\end{figure}

The SketchUp 3D modelling software \citep{trimble2014sketchup} provides fine control of position and orientation while allowing production of realistic images with a variety of persons and scenes. A number of synthetic 3D person models and 3D scenes were obtained from the SketchUp 3D Warehouse \citep{trimble2014warehouse}. Persons were placed randomly in scenes, rotated about the vertical axis, and exposed to various lighting conditions. Figure~\ref{fig:sketchupPersons} shows a subset of the diverse persons and scenes, with random lighting and rotation. The camera was positioned so that it pointed at the person, but at elevations of $0$, $\sfrac{\pi}{8}$, and $\sfrac{\pi}{4}$ radians, as illustrated in Figure~\ref{fig:camera}.

The dataset consists of $3$ instances $\times$ $68$ persons $\times$ $13$ scenes $\times$ $3$ camera elevations $= 7956$ images. These images were cropped and scaled to 64x128, to be consistent with the INRIA Person Dataset \citep{dalal2005histograms}.  Background (person-less) images were obtained from 3D scene models without persons present.

\subsection{Synthetic data results}
A separate linear SVM was trained on HOG features from each of the viewpoints, and each trained classifier was tested on the $\sfrac{\pi}{4}$ viewpoint. Then each classifier was adapted to the $\sfrac{\pi}{4}$ viewpoint according to Algorithm~\ref{alg:adapt}, and re-tested. Note that the transformation from the $\sfrac{\pi}{4}$ training data to the $\sfrac{\pi}{4}$ viewpoint is the identity, so no comparison is necessary. We also reproduced the results of \citet{li2008human} using these data (Algorithm~\ref{alg:li}). We used a modified nine-fold cross-validation scheme ensuring that neither scenes nor persons were shared between the training and testing sets.

\setlength{\figurewidth}{0.5\linewidth}
\begin{figure}[tb]
\begin{center}
	\input{\folder/20170113b_sketchupResults.tex}
\end{center}
   \caption{\label{fig:classifierWarping}
Performance is shown for a single-view person detector trained on images from three elevations ($0$, $\sfrac{\pi}{8}$, and $\sfrac{\pi}{4}$) and tested on images from $\sfrac{\pi}{4}$ elevation (indicated by solid, dashed, and dotted red curves, respectively). Our viewpoint adaptation method is illustrated by the purple curves for each of the mismatched detectors, while the image warping method of \citet{li2008human} is represented by blue curves. The achievable frame rates of these methods, where our approach has the clear advantage, is discussed in Section~\ref{sec:frameRate}.}
\end{figure}
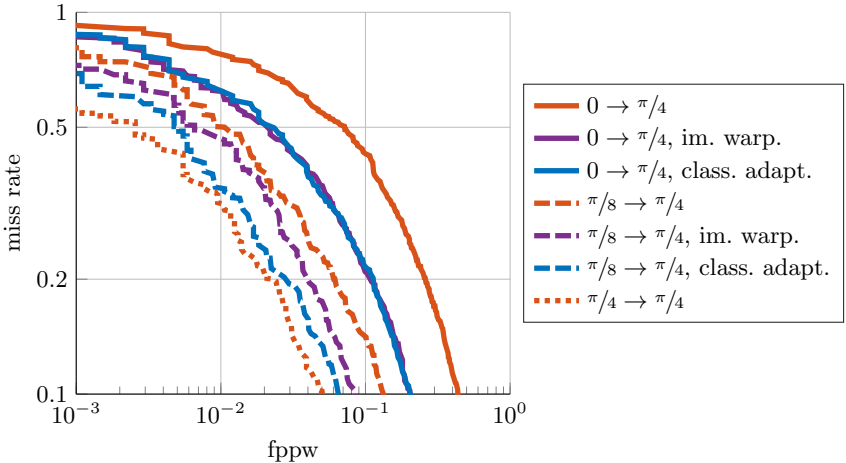

Figure~\ref{fig:classifierWarping} illustrates the performance of each of these detectors. The similarity between the image warping method of \citet{li2008human} and our analogous classifier adaptation is evidence that the feature re-mapping method described above accurately approximates the more computationally expensive image warping. Furthermore, each viewpoint-adapted classifier performs at least as well as the corresponding original classifier, with greater performance gains for source viewpoints farther from the target viewpoint. The main benefit of the proposed approach over that of \citet{li2008human} is in the vastly improved computation time, as we will see in Section~\ref{sec:frameRate}.

One potential failing of the proposed viewpoint adaptation algorithm is its reliance on camera pose information. Such information can be acquired by calibration with a known physical target or utilizing some known scene structure \citep{robertson2009structure}. These methods will lead to varying degrees of inaccuracy in the estimated camera pose. We sought, therefore, to explore the robustness of our algorithm with respect to mis-estimation of the camera's elevation.

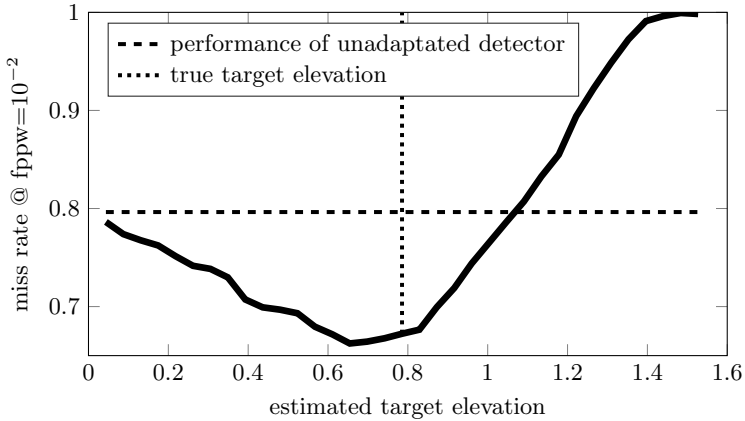
\begin{figure}[tb]
\begin{center}
	\input{\folder/20160308_sketchupRobustness.tex}
\end{center}
   \caption{\label{fig:angleRobustness} A trained $0$-elevation detector was adapted to a range of elevations from $0$ to $\sfrac{\pi}{2}$ radians. It was then evaluated on data from an elevation of $\sfrac{\pi}{4}$. The miss rate at $10^{-2}$ false positives per window is plotted as a function of the elevation to which the detector was adapted. Our approach uniformly improves performance until the estimated target elevation is more than 0.31 radians greater than the true elevation.}
\end{figure}

We used a trained $0$-elevation classifier and attempted to apply it to images from an elevation of $\sfrac{\pi}{4}$ radians. To model our uncertainty in that testing elevation, we adapted the classifier to a range of elevations from $0$ to $\sfrac{\pi}{2}$, applying each to the $\sfrac{\pi}{4}$ data. The results are illustrated in Figure~\ref{fig:angleRobustness}.

\setlength{\figurewidth}{0.7\linewidth}

The best performance, measured by the miss rate at $10^{-2}$ false positives per window (fppw), is achieved when the estimated and true target elevation are approximately equal. Performance decays as the estimated elevation decreases toward $0$ (the source elevation). As the estimated elevation increases beyond $\sfrac{\pi}{4}$, performance decays rapidly, falling below the performance of the original, unadapted detector around $1.1$ radians. This means that, for this scenario, classifier adaptation is advantageous for errors of up to $0.31$ radians $(\approx18^\circ)$ from the true target elevation. This is much greater than the orientation estimation errors of modern camera calibration approaches, which are on the order of $0.2-0.25^\circ$ \citep{antone2000automatic,robertson2009structure}.

\subsection{Runtime analysis}
\label{sec:frameRate}
In analysing the runtime of perspective warping methods, it is important to understand that runtime does not depend only on the frame size (as in most such analyses, e.g. \citet{dollar2014fast}), but also on the perspective of the scene. This is because the number of potential detection windows (computed as described in Section~\ref{sec:windows}) depends on the camera perspective. Thus our analysis looks specifically at the CAVIAR dataset and its associated camera perspective.

\begin{table}[tb]
\begin{center}
\begin{tabular}{|l|l|}
\hline
Method & Frame rate (fps) \\
\hline\hline
HOG & 0.439\\
ACF & 3.54 \\
\hline
HOG im. warp. & 0.0666\\
ACF im. warp. & 0.0419 \\
\hline
HOG class. adapt. & 11.8\\
ACF class. adapt. & 10.7\\
\hline
\end{tabular}
\end{center}
\caption{\label{tab:runtimeResults}Frame rates for each algorithm for the CAVIAR dataset.}
\end{table}

Table~\ref{tab:runtimeResults} shows the frame rates (using a single Intel Core i7 CPU) for each method when applied to images from the CAVIAR dataset. The most important observation to make here is that image warping methods operate at more than $15$ \emph{seconds per frame}, while the classifier adaptation methods are even faster than the unmodified detectors. Image warping methods are very slow because they must compute features for each detection window independently (see Algorithm~\ref{alg:li}). For the CAVIAR dataset, we considered 31603 detection windows, each of which is $128\times64$ pixels, so features must be computed over a total of $2.59\times10^8$ pixels. For comparison, the unmodified ACF detector, applied to the same image, computes features over only $2.36\times10^6$ pixels. Indeed, we see about 2 orders of magnitude difference in the resulting frame rates.

A few other observations can be made about these timing results. First, since the computation time for the viewpoint-informed methods is driven by the feature computation time per pixel (without scaling), the adapted HOG detectors are slightly faster than the corresponding adapted ACF detectors, defying the trend in their unadapted counterparts. Second, to ensure a fair comparison to the adapted detectors, the unadapted detectors upsampled the original $384\times288$ images by a factor of 4. This enabled them to detect small-scale persons but also resulted in frame rates somewhat lower than those advertised in the literature for $640\times480$ images \citep{dollar2014fast}.

The viewpoint-adapted detectors achieve higher frame rates than the unadapted detectors by implicitly using camera pose information to limit the scale-positions at which persons may be detected. The choice of candidate detection windows is discussed in detail in Section~\ref{sec:windows}. This speedup is conceptually similar to that achieved by \citet{benenson2012pedestrian} by using a stereo camera.

\section{Conclusion}
This work is the first to offer a viewpoint adaptation framework limiting the detection search space and handling perspective warping, while retaining frame rates suitable for real-time applications. Our results demonstrate that viewpoint-adapted detectors can outperform even state-of-the-art detectors that lack perspective correction. Our novel detector adaptation method was demonstrated for linear classifiers, using HOG/SVM as a canonical example, and for the more sophisticated ACF detector \citep{dollar2014fast}. Similar derivations may enable viewpoint adaptation of even more diverse object detection schemes like deformable-parts models \citep{felzenszwalb2010object} and convolutional neural networks \citep{krizhevsky2012imagenet}. Such implementations are left as future work.

Our viewpoint adaptation algorithm improves detection performance using no training data from the target viewpoint, requiring only information about the pose of the camera with respect to the ground plane. Furthermore, the detector adaptation algorithm is fairly robust with respect to mis-estimation of the target viewpoint, providing performance improvements even for relatively large errors in the estimated viewpoint. These methods may be extended to explicitly handle uncertainty in the camera pose estimate, as well as to detect objects with more complex 3D models.


\bibliographystyle{splncs}
\bibliography{refs}
\end{document}

%% file: 20160304_imageWarpingExample.tex
%
\begin{tikzpicture}

\begin{axis}[%
width=0.308\figurewidth,
height=0.617\figurewidth,
at={(0\figurewidth,0.648\figurewidth)},
scale only axis,
axis on top,
xmin=0.5,
xmax=64.5,
tick align=outside,
scaled ticks=false,
tick label style={/pgf/number format/fixed},
y dir=reverse,
ymin=0.5,
ymax=128.5,
ylabel={original},
ticks=none,
title={$\theta=0$}
]
\addplot [forget plot] graphics [xmin=0.5,xmax=64.5,ymin=0.5,ymax=128.5] {\folder/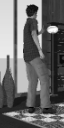};
\end{axis}

\begin{axis}[%
width=0.308\figurewidth,
height=0.617\figurewidth,
at={(0.346\figurewidth,0.648\figurewidth)},
scale only axis,
axis on top,
xmin=0.5,
xmax=64.5,
tick align=outside,
scaled ticks=false,
tick label style={/pgf/number format/fixed},
y dir=reverse,
ymin=0.5,
ymax=128.5,
ticks=none,
title={$\theta=\sfrac{\pi}{8}$}
]
\addplot [forget plot] graphics [xmin=0.5,xmax=64.5,ymin=0.5,ymax=128.5] {\folder/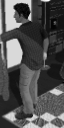};
\end{axis}

\begin{axis}[%
width=0.308\figurewidth,
height=0.617\figurewidth,
at={(0.692\figurewidth,0.648\figurewidth)},
scale only axis,
axis on top,
xmin=0.5,
xmax=64.5,
tick align=outside,
scaled ticks=false,
tick label style={/pgf/number format/fixed},
y dir=reverse,
ymin=0.5,
ymax=128.5,
ticks=none,
title={$\theta=\sfrac{\pi}{4}$}
]
\addplot [forget plot] graphics [xmin=0.5,xmax=64.5,ymin=0.5,ymax=128.5] {\folder/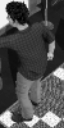};
\end{axis}

\begin{axis}[%
width=0.308\figurewidth,
height=0.617\figurewidth,
at={(0\figurewidth,0\figurewidth)},
scale only axis,
axis on top,
xmin=0.5,
xmax=64.5,
tick align=outside,
scaled ticks=false,
tick label style={/pgf/number format/fixed},
y dir=reverse,
ymin=0.5,
ymax=128.5,
ylabel={warped},
ticks=none
]
\addplot [forget plot] graphics [xmin=0.5,xmax=64.5,ymin=0.5,ymax=128.5] {\folder/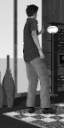};
\end{axis}

\begin{axis}[%
width=0.308\figurewidth,
height=0.617\figurewidth,
at={(0.346\figurewidth,0\figurewidth)},
scale only axis,
axis on top,
xmin=0.5,
xmax=64.5,
tick align=outside,
scaled ticks=false,
tick label style={/pgf/number format/fixed},
y dir=reverse,
ymin=0.5,
ymax=128.5,
ticks=none
]
\addplot [forget plot] graphics [xmin=0.5,xmax=64.5,ymin=0.5,ymax=128.5] {\folder/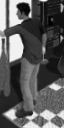};
\end{axis}

\begin{axis}[%
width=0.308\figurewidth,
height=0.617\figurewidth,
at={(0.692\figurewidth,0\figurewidth)},
scale only axis,
axis on top,
xmin=0.5,
xmax=64.5,
tick align=outside,
scaled ticks=false,
tick label style={/pgf/number format/fixed},
y dir=reverse,
ymin=0.5,
ymax=128.5,
ticks=none
]
\addplot [forget plot] graphics [xmin=0.5,xmax=64.5,ymin=0.5,ymax=128.5] {\folder/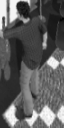};
\end{axis}

\end{tikzpicture}%

%% file: 20160401_adaptationRocsPets.tex
%
%


\begin{axis}[%
width=0.956\figurewidth,
height=0.878\figurewidth,
at={(1\figurewidth,0\figurewidth)},
scale only axis,
xmode=log,
xmin=0.005,
xmax=2,
xminorticks=true,
xlabel={false positives per image},
xmajorgrids,
scaled ticks=false,
tick label style={/pgf/number format/fixed},
ymode=log,
ymin=0.4,
ymax=1,
ytick={0.5, 0.6, 0.7, 0.8, 0.9,   1},
yticklabels={},
yminorticks=true,
ylabel={},
ymajorgrids,
axis background/.style={fill=white},
axis x line*=bottom,
axis y line*=left,
legend style={at={(0,0)},font=\tiny,anchor=south west,legend cell align=left,align=left,draw=white!15!black}
]
\addplot [color=mycolor1,solid,line width=2.0pt]
  table[row sep=crcr]{%
0	0.999809741248097\\
0	0.968797564687976\\
0.000666666666666667	0.964802130898021\\
0.000666666666666667	0.945205479452055\\
0.002	0.935502283105023\\
0.00233333333333333	0.911149162861492\\
0.004	0.901065449010654\\
0.00466666666666667	0.886986301369863\\
0.008	0.876141552511416\\
0.00966666666666667	0.862633181126332\\
0.0103333333333333	0.848744292237443\\
0.012	0.837899543378995\\
0.014	0.834284627092846\\
0.0146666666666667	0.824200913242009\\
0.0166666666666667	0.82058599695586\\
0.0203333333333333	0.802321156773212\\
0.0216666666666667	0.80003805175038\\
0.0226666666666667	0.78234398782344\\
0.0263333333333333	0.767694063926941\\
0.0266666666666667	0.758942161339422\\
0.028	0.758561643835616\\
0.0296666666666667	0.74486301369863\\
0.0326666666666667	0.734398782343988\\
0.0356666666666667	0.72945205479452\\
0.0356666666666667	0.725456621004566\\
0.043	0.704337899543379\\
0.047	0.690829528158295\\
0.0513333333333333	0.688926940639269\\
0.0543333333333333	0.682267884322679\\
0.0573333333333333	0.679794520547945\\
0.0586666666666667	0.67351598173516\\
0.0646666666666667	0.664954337899543\\
0.0723333333333333	0.659436834094368\\
0.073	0.655441400304414\\
0.0793333333333333	0.652777777777778\\
0.0846666666666667	0.64783105022831\\
0.0866666666666667	0.643645357686454\\
0.0953333333333333	0.638508371385084\\
0.101333333333333	0.632039573820396\\
0.106	0.631088280060883\\
0.108666666666667	0.627283105022831\\
0.117	0.623668188736682\\
0.125666666666667	0.621955859969559\\
0.128	0.618531202435312\\
0.136	0.613013698630137\\
0.140666666666667	0.612252663622527\\
0.147666666666667	0.60806697108067\\
0.157666666666667	0.605593607305936\\
0.159333333333333	0.603500761035008\\
0.176333333333333	0.600076103500761\\
0.185333333333333	0.595890410958904\\
0.214333333333333	0.592465753424658\\
0.227	0.587899543378995\\
0.234666666666667	0.587899543378995\\
0.246666666666667	0.584665144596651\\
0.258333333333333	0.582572298325723\\
0.269333333333333	0.58238203957382\\
0.279666666666667	0.581050228310502\\
0.285333333333333	0.578957382039574\\
0.306	0.575913242009132\\
0.326666666666667	0.572108066971081\\
0.346	0.56982496194825\\
0.366	0.568683409436834\\
0.369666666666667	0.567541856925419\\
0.383333333333333	0.566590563165906\\
0.398333333333333	0.564117199391172\\
0.431666666666667	0.562024353120244\\
0.439333333333333	0.560502283105023\\
0.457	0.559170471841705\\
0.48	0.558599695585997\\
0.487	0.556126331811263\\
0.501	0.555745814307458\\
0.506	0.554794520547945\\
0.517666666666667	0.554794520547945\\
0.542666666666667	0.551940639269406\\
0.552	0.551940639269406\\
0.569666666666667	0.550228310502283\\
0.597	0.548706240487062\\
0.608	0.546803652968036\\
0.614333333333333	0.546803652968036\\
0.624333333333333	0.544330289193303\\
0.639666666666667	0.542808219178082\\
0.665	0.541666666666667\\
0.691	0.541476407914764\\
0.699333333333333	0.539764079147641\\
0.723	0.539383561643836\\
0.752333333333333	0.53748097412481\\
0.768333333333333	0.536910197869102\\
0.788333333333333	0.534056316590563\\
0.801333333333333	0.533675799086758\\
0.807333333333333	0.532534246575342\\
0.823666666666667	0.532153729071537\\
0.845	0.530631659056317\\
0.858333333333333	0.530251141552511\\
0.863	0.529109589041096\\
0.888	0.528158295281583\\
0.92	0.527587519025875\\
0.926333333333333	0.526636225266362\\
0.942	0.526255707762557\\
0.942	0.525684931506849\\
0.975333333333333	0.524543378995434\\
1.023	0.523401826484018\\
1.03566666666667	0.522260273972603\\
1.07566666666667	0.519786910197869\\
1.07766666666667	0.519025875190259\\
1.111	0.518074581430746\\
1.12533333333333	0.516362252663622\\
1.14866666666667	0.516362252663622\\
1.18166666666667	0.514649923896499\\
1.218	0.514459665144597\\
1.22433333333333	0.513318112633181\\
1.26266666666667	0.511986301369863\\
1.27666666666667	0.51027397260274\\
1.29566666666667	0.509322678843227\\
1.32133333333333	0.508942161339422\\
1.33366666666667	0.507800608828006\\
1.361	0.507229832572298\\
1.36466666666667	0.506468797564688\\
1.39966666666667	0.50589802130898\\
1.41566666666667	0.50513698630137\\
1.46	0.50437595129376\\
1.49633333333333	0.502473363774734\\
1.571	0.501141552511416\\
1.573	0.500380517503805\\
1.602	0.500190258751903\\
1.61266666666667	0.49923896499239\\
1.63733333333333	0.49923896499239\\
1.63733333333333	0.498858447488584\\
1.66133333333333	0.498858447488584\\
1.68833333333333	0.497907153729072\\
1.72466666666667	0.497146118721461\\
1.75033333333333	0.496955859969559\\
1.78333333333333	0.496004566210046\\
1.81733333333333	0.495814307458143\\
1.83166666666667	0.49486301369863\\
1.854	0.49486301369863\\
1.86633333333333	0.49410197869102\\
1.89366666666667	0.493721461187215\\
1.91166666666667	0.491818873668189\\
1.94333333333333	0.491248097412481\\
1.94333333333333	0.490867579908676\\
1.96466666666667	0.489345509893455\\
1.98333333333333	0.489155251141553\\
2.022	0.487633181126332\\
2.02666666666667	0.486491628614916\\
2.04666666666667	0.486301369863014\\
2.04666666666667	0.485920852359209\\
2.09166666666667	0.485350076103501\\
2.09366666666667	0.484779299847793\\
2.13166666666667	0.484208523592085\\
2.164	0.481925418569254\\
2.18033333333333	0.481925418569254\\
2.19166666666667	0.480593607305936\\
2.20933333333333	0.480022831050228\\
2.25133333333333	0.479452054794521\\
2.25533333333333	0.47869101978691\\
2.275	0.478310502283105\\
2.291	0.477168949771689\\
2.30866666666667	0.476978691019787\\
2.356	0.475076103500761\\
2.356	0.474695585996956\\
2.40433333333333	0.474124809741248\\
2.414	0.473363774733638\\
2.439	0.472602739726027\\
2.49433333333333	0.471461187214612\\
2.495	0.470890410958904\\
2.522	0.470890410958904\\
2.542	0.470319634703196\\
2.56033333333333	0.469939117199391\\
2.56533333333333	0.469178082191781\\
2.59833333333333	0.46841704718417\\
2.61333333333333	0.467085235920852\\
2.63733333333333	0.46689497716895\\
2.64866666666667	0.466133942161339\\
2.68966666666667	0.465943683409437\\
2.72033333333333	0.465182648401826\\
2.72233333333333	0.464421613394216\\
2.75733333333333	0.463660578386606\\
2.79266666666667	0.463660578386606\\
2.817	0.462328767123288\\
2.836	0.462328767123288\\
2.86166666666667	0.460616438356164\\
2.866	0.459665144596651\\
2.88933333333333	0.459094368340944\\
2.926	0.457191780821918\\
2.94466666666667	0.457191780821918\\
2.99	0.455669710806697\\
3.01633333333333	0.455479452054795\\
3.018	0.454908675799087\\
3.05366666666667	0.454718417047184\\
3.061	0.454337899543379\\
3.13866666666667	0.453386605783866\\
3.17933333333333	0.452625570776256\\
3.197	0.451864535768645\\
3.20366666666667	0.450913242009132\\
3.22	0.450913242009132\\
3.24	0.449581430745814\\
3.258	0.449581430745814\\
3.28133333333333	0.448630136986301\\
3.29266666666667	0.447488584474886\\
};
\addlegendentry{(65.5\%)};

\addplot [color=mycolor2,solid,line width=2.0pt]
  table[row sep=crcr]{%
0	0.999809741248097\\
0	0.923325722983257\\
0.001	0.853500761035008\\
0.00233333333333333	0.847983257229833\\
0.00233333333333333	0.81544901065449\\
0.00466666666666667	0.79927701674277\\
0.007	0.763508371385084\\
0.0113333333333333	0.74923896499239\\
0.015	0.717085235920852\\
0.0213333333333333	0.693493150684932\\
0.0256666666666667	0.687595129375951\\
0.0373333333333333	0.658485540334855\\
0.0596666666666667	0.623858447488584\\
0.0746666666666667	0.611301369863014\\
0.0843333333333333	0.607496194824962\\
0.103	0.593607305936073\\
0.159333333333333	0.571537290715373\\
0.181333333333333	0.565068493150685\\
0.211666666666667	0.559360730593607\\
0.217	0.554984779299848\\
0.231	0.554033485540335\\
0.259333333333333	0.547945205479452\\
0.283	0.53824200913242\\
0.307666666666667	0.534627092846271\\
0.353333333333333	0.529680365296804\\
0.359666666666667	0.526636225266362\\
0.377666666666667	0.524923896499239\\
0.412	0.517503805175038\\
0.468	0.511415525114155\\
0.481666666666667	0.508561643835616\\
0.555	0.502853881278539\\
0.572666666666667	0.499809741248097\\
0.616333333333333	0.496575342465753\\
0.641333333333333	0.493531202435312\\
0.647	0.491057838660578\\
0.705666666666667	0.485350076103501\\
0.720666666666667	0.482305936073059\\
0.751	0.479832572298326\\
0.817666666666667	0.476027397260274\\
0.826666666666667	0.474315068493151\\
0.874333333333333	0.470700152207002\\
0.896666666666667	0.465753424657534\\
0.929666666666667	0.464041095890411\\
0.976	0.457762557077626\\
0.995333333333333	0.457191780821918\\
1.036	0.452435312024353\\
1.061	0.447488584474886\\
1.093	0.44482496194825\\
1.14633333333333	0.441780821917808\\
1.17133333333333	0.437975646879756\\
1.20733333333333	0.436073059360731\\
1.285	0.42865296803653\\
1.34633333333333	0.425608828006088\\
1.46033333333333	0.416666666666667\\
1.47766666666667	0.414383561643836\\
1.502	0.413812785388128\\
1.52366666666667	0.411529680365297\\
1.56	0.409817351598174\\
1.59966666666667	0.404870624048706\\
1.623	0.404109589041096\\
1.62666666666667	0.402016742770167\\
1.65533333333333	0.399353120243531\\
1.707	0.397260273972603\\
1.73266666666667	0.392884322678843\\
1.79433333333333	0.389649923896499\\
1.809	0.387366818873668\\
1.83766666666667	0.38679604261796\\
1.86966666666667	0.382990867579909\\
1.91166666666667	0.38013698630137\\
1.995	0.37576103500761\\
2.02133333333333	0.375570776255708\\
2.07833333333333	0.369292237442922\\
2.16033333333333	0.365867579908676\\
2.191	0.362633181126332\\
2.237	0.361681887366819\\
2.24866666666667	0.359779299847793\\
2.31266666666667	0.356164383561644\\
2.33666666666667	0.353120243531202\\
2.38	0.350837138508371\\
2.458	0.347983257229833\\
2.52633333333333	0.343226788432268\\
2.60533333333333	0.338660578386606\\
2.674	0.33751902587519\\
2.708	0.335806697108067\\
2.76133333333333	0.3308599695586\\
2.79933333333333	0.328957382039574\\
2.85133333333333	0.327625570776256\\
2.91733333333333	0.322678843226788\\
2.98033333333333	0.318873668188737\\
3.156	0.312404870624049\\
3.32266666666667	0.308409436834094\\
3.33466666666667	0.307458143074581\\
3.39666666666667	0.306887366818874\\
3.40733333333333	0.305555555555556\\
3.548	0.299847792998478\\
3.58733333333333	0.297184170471842\\
3.64533333333333	0.296232876712329\\
3.65833333333333	0.295281582952816\\
3.737	0.293949771689498\\
3.742	0.292998477929985\\
3.80233333333333	0.290715372907154\\
3.832	0.290715372907154\\
3.91433333333333	0.287861491628615\\
3.98533333333333	0.286910197869102\\
4.089	0.283675799086758\\
4.212	0.281202435312024\\
4.22133333333333	0.279870624048706\\
4.276	0.279490106544901\\
4.325	0.278348554033486\\
4.338	0.27720700152207\\
4.448	0.274923896499239\\
4.48466666666667	0.273211567732116\\
4.531	0.273021308980213\\
4.599	0.271118721461187\\
4.60633333333333	0.269977168949772\\
4.66033333333333	0.269216133942161\\
4.70766666666667	0.267123287671233\\
4.749	0.26693302891933\\
4.8	0.26541095890411\\
4.86166666666667	0.262747336377473\\
4.92133333333333	0.262557077625571\\
4.97566666666667	0.261415525114155\\
5.176	0.258561643835616\\
5.21633333333333	0.256468797564688\\
5.275	0.254566210045662\\
5.32566666666667	0.254566210045662\\
5.49833333333333	0.249809741248097\\
5.546	0.249809741248097\\
5.67133333333333	0.245433789954338\\
5.73733333333333	0.244482496194825\\
5.801	0.244292237442922\\
5.81333333333333	0.242960426179604\\
5.84966666666667	0.242960426179604\\
5.86033333333333	0.241628614916286\\
5.89566666666667	0.241248097412481\\
5.91933333333333	0.23972602739726\\
6.02766666666667	0.238394216133942\\
6.05066666666667	0.237252663622527\\
6.10733333333333	0.236491628614916\\
6.19466666666667	0.23306697108067\\
6.23666666666667	0.232876712328767\\
6.25966666666667	0.231735159817352\\
6.363	0.230783866057839\\
6.46166666666667	0.227739726027397\\
6.59233333333333	0.225076103500761\\
6.656	0.224695585996956\\
6.67733333333333	0.22279299847793\\
6.70733333333333	0.22279299847793\\
6.81666666666667	0.219939117199391\\
6.87133333333333	0.219939117199391\\
6.93333333333333	0.218607305936073\\
6.95133333333333	0.217085235920852\\
6.98733333333333	0.21689497716895\\
7.01833333333333	0.215182648401826\\
7.08966666666667	0.213660578386606\\
7.15033333333333	0.213280060882801\\
7.156	0.21251902587519\\
7.238	0.212328767123288\\
7.308	0.210806697108067\\
7.34933333333333	0.210616438356164\\
7.45833333333333	0.207191780821918\\
7.55766666666667	0.204908675799087\\
7.64633333333333	0.204337899543379\\
7.686	0.202625570776256\\
7.75933333333333	0.201864535768645\\
7.88333333333333	0.199391171993912\\
8.088	0.197488584474886\\
8.13933333333333	0.19558599695586\\
8.25666666666667	0.193493150684932\\
8.30466666666667	0.193112633181126\\
8.42	0.190068493150685\\
8.48866666666667	0.190068493150685\\
8.51966666666667	0.188926940639269\\
8.62833333333333	0.188165905631659\\
8.67066666666667	0.186834094368341\\
8.759	0.186073059360731\\
8.78966666666667	0.185121765601218\\
8.91066666666667	0.18455098934551\\
8.93766666666667	0.183219178082192\\
9.15666666666667	0.181697108066971\\
9.34033333333333	0.178082191780822\\
9.53633333333333	0.175989345509893\\
9.59533333333333	0.175989345509893\\
9.69766666666667	0.175228310502283\\
9.72266666666667	0.174467275494673\\
9.94966666666667	0.173325722983257\\
10.0103333333333	0.172564687975647\\
10.087	0.172564687975647\\
10.1153333333333	0.171803652968037\\
10.2816666666667	0.171042617960426\\
10.352	0.170471841704718\\
10.3896666666667	0.1691400304414\\
10.514	0.1691400304414\\
10.6393333333333	0.168569254185693\\
10.6823333333333	0.166856925418569\\
10.8696666666667	0.166286149162861\\
10.92	0.165334855403349\\
10.9816666666667	0.165334855403349\\
11.0743333333333	0.163812785388128\\
11.1256666666667	0.163812785388128\\
};
\addlegendentry{(59.4\%)};

\addplot [color=mycolor3,dashed,line width=2.0pt]
  table[row sep=crcr]{%
0	0.999809741248097\\
0.000666666666666667	0.97203196347032\\
0.00333333333333333	0.96175799086758\\
0.004	0.927321156773212\\
0.00566666666666667	0.899923896499239\\
0.00566666666666667	0.844368340943683\\
0.0113333333333333	0.781202435312024\\
0.0126666666666667	0.779299847792998\\
0.0146666666666667	0.754185692541857\\
0.0186666666666667	0.741628614916286\\
0.02	0.724315068493151\\
0.0236666666666667	0.715943683409437\\
0.0243333333333333	0.706811263318113\\
0.0273333333333333	0.70148401826484\\
0.03	0.686643835616438\\
0.0393333333333333	0.670091324200913\\
0.0446666666666667	0.65658295281583\\
0.0553333333333333	0.646118721461187\\
0.0646666666666667	0.636986301369863\\
0.07	0.627663622526636\\
0.083	0.617389649923896\\
0.0863333333333333	0.611872146118722\\
0.0933333333333333	0.606354642313546\\
0.108666666666667	0.600266362252664\\
0.116	0.592846270928463\\
0.123666666666667	0.590943683409437\\
0.13	0.586377473363775\\
0.143666666666667	0.580098934550989\\
0.153666666666667	0.572298325722983\\
0.164333333333333	0.57058599695586\\
0.174666666666667	0.564878234398782\\
0.180666666666667	0.563736681887367\\
0.188666666666667	0.558980213089802\\
0.213	0.554604261796043\\
0.237666666666667	0.547564687975647\\
0.241666666666667	0.543759512937595\\
0.266333333333333	0.536910197869102\\
0.273666666666667	0.531773211567732\\
0.282333333333333	0.531202435312024\\
0.311333333333333	0.520167427701674\\
0.332	0.513888888888889\\
0.353	0.509893455098935\\
0.372	0.503234398782344\\
0.385666666666667	0.499619482496195\\
0.403	0.497336377473364\\
0.416333333333333	0.493531202435312\\
0.437	0.489916286149163\\
0.442333333333333	0.486301369863014\\
0.459	0.482876712328767\\
0.466333333333333	0.482876712328767\\
0.496666666666667	0.477168949771689\\
0.496666666666667	0.476217656012177\\
0.513	0.475266362252664\\
0.525	0.473363774733638\\
0.538666666666667	0.467465753424658\\
0.544	0.467085235920852\\
0.558	0.461377473363775\\
0.570666666666667	0.459474885844749\\
0.586333333333333	0.455479452054795\\
0.597333333333333	0.455098934550989\\
0.615666666666667	0.452054794520548\\
0.652	0.448439878234399\\
0.656	0.446917808219178\\
0.681	0.445015220700152\\
0.704666666666667	0.4412100456621\\
0.711333333333333	0.439117199391172\\
0.727	0.437975646879756\\
0.730333333333333	0.436453576864536\\
0.749666666666667	0.433980213089802\\
0.769	0.433409436834094\\
0.789666666666667	0.42865296803653\\
0.81	0.428082191780822\\
0.813333333333333	0.426560121765601\\
0.854	0.421803652968037\\
0.865333333333333	0.4191400304414\\
0.903	0.416666666666667\\
0.907666666666667	0.414954337899543\\
0.932333333333333	0.412861491628615\\
0.939	0.410958904109589\\
0.967666666666667	0.40734398782344\\
0.994	0.405821917808219\\
1.01633333333333	0.402016742770167\\
1.03233333333333	0.400494672754947\\
1.038	0.398592085235921\\
1.06066666666667	0.397450532724505\\
1.06733333333333	0.395928462709285\\
1.08633333333333	0.394977168949772\\
1.09333333333333	0.393264840182648\\
1.11466666666667	0.39117199391172\\
1.15733333333333	0.389079147640791\\
1.16233333333333	0.387557077625571\\
1.17933333333333	0.386986301369863\\
1.215	0.382229832572298\\
1.26466666666667	0.379566210045662\\
1.29433333333333	0.376712328767123\\
1.334	0.37423896499239\\
1.34866666666667	0.372716894977169\\
1.364	0.372716894977169\\
1.40066666666667	0.369482496194825\\
1.40733333333333	0.367960426179604\\
1.44566666666667	0.365867579908676\\
1.46266666666667	0.363584474885845\\
1.49033333333333	0.361491628614916\\
1.518	0.361111111111111\\
1.528	0.359969558599696\\
1.584	0.358257229832572\\
1.64	0.354832572298326\\
1.669	0.354071537290715\\
1.67066666666667	0.3529299847793\\
1.709	0.351407914764079\\
1.717	0.350266362252664\\
1.76733333333333	0.348363774733638\\
1.78166666666667	0.346461187214612\\
1.82666666666667	0.344558599695586\\
1.84933333333333	0.34265601217656\\
1.85933333333333	0.340943683409437\\
1.888	0.339992389649924\\
1.903	0.338660578386606\\
1.92433333333333	0.338660578386606\\
1.93966666666667	0.33751902587519\\
2.00066666666667	0.336948249619482\\
2.01566666666667	0.335616438356164\\
2.04133333333333	0.335235920852359\\
2.07266666666667	0.333143074581431\\
2.127	0.331240487062405\\
2.168	0.328767123287671\\
2.18866666666667	0.328767123287671\\
2.197	0.327625570776256\\
2.248	0.32648401826484\\
2.25533333333333	0.325342465753425\\
2.308	0.324391171993912\\
2.329	0.323059360730594\\
2.35266666666667	0.323059360730594\\
2.43033333333333	0.318683409436834\\
2.476	0.317922374429224\\
2.48833333333333	0.317161339421613\\
2.53666666666667	0.316400304414003\\
2.57333333333333	0.313926940639269\\
2.6	0.313165905631659\\
2.63533333333333	0.313165905631659\\
2.685	0.312214611872146\\
2.714	0.31031202435312\\
2.76866666666667	0.3087899543379\\
2.771	0.308219178082192\\
2.82633333333333	0.307648401826484\\
2.87366666666667	0.305365296803653\\
2.882	0.304223744292237\\
2.946	0.302130898021309\\
2.966	0.300418569254186\\
2.994	0.300038051750381\\
3.014	0.298325722983257\\
3.02966666666667	0.298325722983257\\
3.04166666666667	0.296993911719939\\
3.09233333333333	0.295662100456621\\
3.125	0.295471841704718\\
3.12733333333333	0.294520547945205\\
3.161	0.2941400304414\\
3.24666666666667	0.290715372907154\\
3.35	0.288622526636225\\
3.36933333333333	0.287671232876712\\
3.39333333333333	0.285388127853881\\
3.46433333333333	0.284436834094368\\
3.51566666666667	0.28310502283105\\
3.54	0.281963470319635\\
3.581	0.28158295281583\\
3.64066666666667	0.279299847792998\\
3.68466666666667	0.278158295281583\\
3.73033333333333	0.277777777777778\\
3.745	0.277016742770167\\
3.79833333333333	0.276826484018265\\
3.802	0.276065449010654\\
3.82966666666667	0.275875190258752\\
3.834	0.275114155251142\\
3.87166666666667	0.273972602739726\\
3.89433333333333	0.272450532724505\\
3.94233333333333	0.272260273972603\\
3.95166666666667	0.271499238964992\\
3.997	0.270928462709285\\
3.997	0.270547945205479\\
4.05033333333333	0.270167427701674\\
4.09766666666667	0.268835616438356\\
4.146	0.268645357686454\\
4.158	0.267884322678843\\
4.218	0.267503805175038\\
4.22733333333333	0.26693302891933\\
4.32566666666667	0.266742770167428\\
4.329	0.265981735159817\\
4.36533333333333	0.265220700152207\\
4.45466666666667	0.264079147640791\\
4.46133333333333	0.263508371385084\\
4.50833333333333	0.262937595129376\\
4.51266666666667	0.262176560121766\\
4.54633333333333	0.261605783866058\\
4.55266666666667	0.260844748858447\\
4.58833333333333	0.260844748858447\\
4.62766666666667	0.260083713850837\\
4.69833333333333	0.260083713850837\\
4.756	0.258942161339422\\
4.80733333333333	0.258942161339422\\
4.813	0.258371385083714\\
};
\addlegendentry{(59.0\%)};

\addplot [color=mycolor4,dashed,line width=2.0pt]
  table[row sep=crcr]{%
0	0.999809741248097\\
0.00133333333333333	0.975076103500761\\
0.003	0.97203196347032\\
0.00366666666666667	0.937785388127854\\
0.00666666666666667	0.925989345509893\\
0.00933333333333333	0.902777777777778\\
0.012	0.899162861491629\\
0.02	0.853500761035008\\
0.021	0.853500761035008\\
0.0253333333333333	0.833523592085236\\
0.0256666666666667	0.82572298325723\\
0.0293333333333333	0.820395738203957\\
0.0346666666666667	0.798135464231355\\
0.0366666666666667	0.784246575342466\\
0.045	0.769596651445966\\
0.0473333333333333	0.760654490106545\\
0.057	0.749619482496195\\
0.0573333333333333	0.741438356164384\\
0.0676666666666667	0.720700152207002\\
0.0716666666666667	0.706430745814308\\
0.0746666666666667	0.705098934550989\\
0.0796666666666667	0.694634703196347\\
0.087	0.686073059360731\\
0.089	0.678082191780822\\
0.0996666666666667	0.664954337899543\\
0.11	0.65810502283105\\
0.118	0.64554794520548\\
0.121333333333333	0.644786910197869\\
0.124666666666667	0.635654490106545\\
0.138666666666667	0.625\\
0.146	0.617389649923896\\
0.160666666666667	0.611872146118722\\
0.165333333333333	0.606735159817352\\
0.169666666666667	0.606164383561644\\
0.178	0.599695585996956\\
0.187333333333333	0.597222222222222\\
0.197666666666667	0.589231354642314\\
0.198666666666667	0.585806697108067\\
0.208	0.578767123287671\\
0.215333333333333	0.57572298325723\\
0.233666666666667	0.564497716894977\\
0.243	0.561073059360731\\
0.250333333333333	0.555936073059361\\
0.263666666666667	0.552511415525114\\
0.271	0.546613394216134\\
0.294333333333333	0.53900304414003\\
0.294666666666667	0.536719939117199\\
0.308333333333333	0.529109589041096\\
0.317333333333333	0.527777777777778\\
0.332333333333333	0.519786910197869\\
0.335	0.519786910197869\\
0.365666666666667	0.511986301369863\\
0.374666666666667	0.507420091324201\\
0.391666666666667	0.503044140030441\\
0.396333333333333	0.499809741248097\\
0.415666666666667	0.495814307458143\\
0.424666666666667	0.492770167427702\\
0.439666666666667	0.490677321156773\\
0.443333333333333	0.487252663622527\\
0.453	0.48458904109589\\
0.469666666666667	0.482305936073059\\
0.469666666666667	0.481544901065449\\
0.488333333333333	0.477359208523592\\
0.498666666666667	0.473934550989346\\
0.512333333333333	0.47279299847793\\
0.534666666666667	0.468797564687976\\
0.538666666666667	0.465372907153729\\
0.555	0.461377473363775\\
0.598333333333333	0.454908675799087\\
0.614333333333333	0.454147640791476\\
0.646	0.45072298325723\\
0.685333333333333	0.443493150684932\\
0.703666666666667	0.442922374429224\\
0.725666666666667	0.4412100456621\\
0.747	0.436453576864536\\
0.770666666666667	0.433219178082192\\
0.797333333333333	0.431126331811263\\
0.818	0.427321156773212\\
0.840333333333333	0.425989345509893\\
0.868666666666667	0.422945205479452\\
0.885333333333333	0.421803652968037\\
0.901666666666667	0.418759512937595\\
0.911	0.418759512937595\\
0.946	0.415334855403349\\
0.968	0.411529680365297\\
0.983	0.410768645357686\\
0.994	0.408295281582953\\
1.04366666666667	0.403538812785388\\
1.057	0.403348554033486\\
1.065	0.401636225266362\\
1.08733333333333	0.399923896499239\\
1.11666666666667	0.398592085235921\\
1.161	0.394406392694064\\
1.177	0.394216133942161\\
1.20333333333333	0.392123287671233\\
1.23266666666667	0.390981735159817\\
1.26233333333333	0.388127853881279\\
1.29566666666667	0.386605783866058\\
1.30766666666667	0.386605783866058\\
1.344	0.383751902587519\\
1.35033333333333	0.382420091324201\\
1.388	0.38089802130898\\
1.44533333333333	0.376712328767123\\
1.46466666666667	0.376331811263318\\
1.49666666666667	0.373097412480974\\
1.52533333333333	0.372146118721461\\
1.53933333333333	0.370433789954338\\
1.552	0.370433789954338\\
1.56933333333333	0.366438356164384\\
1.60433333333333	0.365106544901065\\
1.63066666666667	0.363013698630137\\
1.644	0.363013698630137\\
1.658	0.361491628614916\\
1.67833333333333	0.360730593607306\\
1.695	0.359208523592085\\
1.72833333333333	0.357876712328767\\
1.748	0.357876712328767\\
1.77766666666667	0.355974124809741\\
1.78333333333333	0.354642313546423\\
1.83066666666667	0.352549467275495\\
1.85	0.352549467275495\\
1.85033333333333	0.351598173515982\\
1.91233333333333	0.34855403348554\\
1.93633333333333	0.34855403348554\\
1.94033333333333	0.347602739726027\\
1.96166666666667	0.347602739726027\\
1.99766666666667	0.345890410958904\\
2.03666666666667	0.343226788432268\\
2.064	0.34265601217656\\
2.099	0.340943683409437\\
2.10266666666667	0.339992389649924\\
2.154	0.338089802130898\\
2.16733333333333	0.336377473363775\\
2.193	0.33599695585997\\
2.22233333333333	0.334665144596651\\
2.278	0.333904109589041\\
2.30533333333333	0.331430745814307\\
2.34366666666667	0.3308599695586\\
2.35666666666667	0.329718417047184\\
2.38033333333333	0.329147640791476\\
2.403	0.327815829528158\\
2.43633333333333	0.327815829528158\\
2.46366666666667	0.325532724505327\\
2.485	0.325532724505327\\
2.51333333333333	0.324010654490107\\
2.549	0.323439878234399\\
2.56466666666667	0.322298325722983\\
2.604	0.321537290715373\\
2.61333333333333	0.320395738203957\\
2.66	0.319063926940639\\
2.68033333333333	0.317732115677321\\
2.694	0.317732115677321\\
2.70833333333333	0.316400304414003\\
2.80133333333333	0.313356164383562\\
2.83033333333333	0.313356164383562\\
2.86033333333333	0.312595129375951\\
2.869	0.311263318112633\\
2.91033333333333	0.31031202435312\\
2.93933333333333	0.308028919330289\\
2.95833333333333	0.307838660578387\\
2.977	0.306506849315068\\
3.04733333333333	0.305936073059361\\
3.05333333333333	0.305365296803653\\
3.092	0.304984779299848\\
3.092	0.304604261796043\\
3.122	0.30441400304414\\
3.14733333333333	0.303462709284627\\
3.18966666666667	0.303272450532725\\
3.19633333333333	0.302511415525114\\
3.22633333333333	0.302321156773212\\
3.26233333333333	0.299847792998478\\
3.28033333333333	0.299847792998478\\
3.29466666666667	0.29851598173516\\
3.31633333333333	0.297754946727549\\
3.399	0.296423135464231\\
3.42033333333333	0.294901065449011\\
3.448	0.294710806697108\\
3.46866666666667	0.293188736681887\\
3.49466666666667	0.292998477929985\\
3.535	0.290334855403349\\
3.56966666666667	0.289383561643836\\
3.59933333333333	0.288622526636225\\
3.678	0.287671232876712\\
3.684	0.287100456621005\\
3.72533333333333	0.286719939117199\\
3.729	0.286149162861492\\
3.80133333333333	0.285197869101979\\
3.831	0.284056316590563\\
3.84066666666667	0.282724505327245\\
3.89633333333333	0.281773211567732\\
3.95166666666667	0.279870624048706\\
3.998	0.279680365296804\\
4.024	0.277777777777778\\
4.04266666666667	0.277587519025875\\
4.065	0.276065449010654\\
4.102	0.274733637747336\\
4.134	0.274733637747336\\
4.17666666666667	0.273782343987823\\
4.19466666666667	0.272450532724505\\
4.24133333333333	0.2720700152207\\
};
\addlegendentry{(64.3\%)};

\addplot [color=mycolor3,dotted,line width=2.0pt]
  table[row sep=crcr]{%
0	0.999809741248097\\
0.000333333333333333	0.980022831050228\\
0.00133333333333333	0.976788432267884\\
0.002	0.961948249619482\\
0.002	0.947108066971081\\
0.00233333333333333	0.947108066971081\\
0.00233333333333333	0.925799086757991\\
0.003	0.922184170471842\\
0.00333333333333333	0.905631659056317\\
0.00433333333333333	0.901065449010654\\
0.00433333333333333	0.877473363774734\\
0.005	0.873477929984779\\
0.005	0.855022831050228\\
0.006	0.840943683409437\\
0.006	0.832191780821918\\
0.007	0.829908675799087\\
0.008	0.819254185692542\\
0.008	0.800799086757991\\
0.00933333333333333	0.797564687975647\\
0.00966666666666667	0.786719939117199\\
0.0106666666666667	0.784817351598174\\
0.0113333333333333	0.769786910197869\\
0.012	0.769216133942161\\
0.0123333333333333	0.758181126331811\\
0.013	0.757229832572298\\
0.013	0.746955859969559\\
0.0143333333333333	0.742009132420091\\
0.0146666666666667	0.733447488584475\\
0.0166666666666667	0.725646879756469\\
0.018	0.71099695585997\\
0.019	0.710616438356164\\
0.02	0.701293759512938\\
0.02	0.692922374429224\\
0.021	0.68531202435312\\
0.0233333333333333	0.683028919330289\\
0.026	0.67275494672755\\
0.0283333333333333	0.670852359208524\\
0.03	0.664954337899543\\
0.0313333333333333	0.664954337899543\\
0.0356666666666667	0.653348554033486\\
0.0386666666666667	0.649353120243531\\
0.045	0.643835616438356\\
0.0476666666666667	0.63679604261796\\
0.0513333333333333	0.634893455098935\\
0.0543333333333333	0.63089802130898\\
0.057	0.63013698630137\\
0.0603333333333333	0.622716894977169\\
0.0673333333333333	0.617579908675799\\
0.0696666666666667	0.613774733637747\\
0.0743333333333333	0.612823439878234\\
0.076	0.608447488584475\\
0.082	0.60445205479452\\
0.0863333333333333	0.597222222222222\\
0.098	0.589421613394216\\
0.105333333333333	0.586377473363775\\
0.109666666666667	0.585806697108067\\
0.116	0.582572298325723\\
0.116333333333333	0.580479452054794\\
0.121	0.577625570776256\\
0.13	0.575913242009132\\
0.135333333333333	0.573439878234399\\
0.145	0.571917808219178\\
0.148333333333333	0.57058599695586\\
0.151	0.566780821917808\\
0.163	0.563926940639269\\
0.170333333333333	0.560692541856925\\
0.172666666666667	0.560692541856925\\
0.183	0.556697108066971\\
0.191666666666667	0.555555555555556\\
0.195666666666667	0.553272450532724\\
0.209333333333333	0.550799086757991\\
0.222333333333333	0.549467275494673\\
0.228666666666667	0.548135464231355\\
0.238333333333333	0.544520547945206\\
0.246333333333333	0.543759512937595\\
0.249	0.542237442922374\\
0.257666666666667	0.540715372907154\\
0.262666666666667	0.537100456621005\\
0.273333333333333	0.536149162861492\\
0.274666666666667	0.535197869101979\\
0.291	0.53310502283105\\
0.291	0.532153729071537\\
0.302333333333333	0.530441400304414\\
0.305666666666667	0.529299847792998\\
0.314	0.529109589041096\\
0.317666666666667	0.527587519025875\\
0.338666666666667	0.524923896499239\\
0.345	0.522640791476408\\
0.363	0.521118721461187\\
0.369333333333333	0.519786910197869\\
0.375	0.517313546423136\\
0.380333333333333	0.517313546423136\\
0.386	0.515220700152207\\
0.403666666666667	0.514459665144597\\
0.410333333333333	0.513318112633181\\
0.413	0.51179604261796\\
0.425333333333333	0.508942161339422\\
0.449	0.507800608828006\\
0.453	0.505707762557078\\
0.458666666666667	0.505327245053272\\
0.465	0.503234398782344\\
0.474333333333333	0.502283105022831\\
0.484333333333333	0.499429223744292\\
0.511333333333333	0.496194824961948\\
0.515333333333333	0.49410197869102\\
0.544333333333333	0.492389649923896\\
0.547	0.491248097412481\\
0.555	0.491057838660578\\
0.560333333333333	0.48896499238965\\
0.577666666666667	0.48820395738204\\
0.580666666666667	0.486491628614916\\
0.592333333333333	0.484779299847793\\
0.609	0.483447488584475\\
0.634333333333333	0.482305936073059\\
0.635333333333333	0.481354642313546\\
0.66	0.479642313546423\\
0.661333333333333	0.478881278538813\\
0.672333333333333	0.478310502283105\\
0.681333333333333	0.476407914764079\\
0.712666666666667	0.473744292237443\\
0.719	0.471270928462709\\
0.727666666666667	0.470509893455099\\
0.754666666666667	0.469178082191781\\
0.758666666666667	0.468036529680365\\
0.770666666666667	0.46689497716895\\
0.771	0.466133942161339\\
0.78	0.464992389649924\\
0.798333333333333	0.464421613394216\\
0.803	0.463470319634703\\
0.813333333333333	0.463470319634703\\
0.819	0.462709284627093\\
0.837	0.46251902587519\\
0.855	0.46099695585997\\
0.858	0.459665144596651\\
0.867333333333333	0.458333333333333\\
0.892666666666667	0.456811263318113\\
0.900333333333333	0.455669710806697\\
0.927666666666667	0.454528158295282\\
0.93	0.453576864535769\\
0.938333333333333	0.453196347031963\\
0.960666666666667	0.453196347031963\\
0.985666666666667	0.452245053272451\\
0.987	0.45148401826484\\
1.00866666666667	0.450342465753425\\
1.02	0.448439878234399\\
1.04233333333333	0.447678843226788\\
1.05266666666667	0.446727549467275\\
1.07	0.446537290715373\\
1.07066666666667	0.445966514459665\\
1.08766666666667	0.44558599695586\\
1.09933333333333	0.44482496194825\\
1.113	0.44482496194825\\
1.13266666666667	0.443302891933029\\
1.15833333333333	0.442732115677321\\
1.15833333333333	0.442351598173516\\
1.176	0.441590563165906\\
1.185	0.440639269406393\\
1.19966666666667	0.44044901065449\\
1.21933333333333	0.438165905631659\\
1.232	0.437785388127854\\
1.23466666666667	0.436834094368341\\
1.261	0.436453576864536\\
1.27266666666667	0.435502283105023\\
1.28666666666667	0.43531202435312\\
1.298	0.434170471841705\\
1.32033333333333	0.434170471841705\\
1.32033333333333	0.4337899543379\\
1.34533333333333	0.433219178082192\\
1.35633333333333	0.431697108066971\\
1.37133333333333	0.431697108066971\\
1.39033333333333	0.430365296803653\\
1.401	0.430365296803653\\
1.415	0.42941400304414\\
1.419	0.428462709284627\\
1.43733333333333	0.427891933028919\\
1.455	0.427891933028919\\
1.462	0.426940639269406\\
1.48066666666667	0.426560121765601\\
1.50366666666667	0.425228310502283\\
1.51966666666667	0.425228310502283\\
1.53466666666667	0.423706240487062\\
1.57566666666667	0.423135464231355\\
1.587	0.422184170471842\\
1.60033333333333	0.421993911719939\\
1.60566666666667	0.421042617960426\\
1.63366666666667	0.420281582952816\\
1.645	0.4191400304414\\
1.668	0.4191400304414\\
1.684	0.418569254185693\\
1.69166666666667	0.417808219178082\\
1.727	0.417237442922374\\
1.74166666666667	0.416666666666667\\
1.76466666666667	0.414764079147641\\
1.77733333333333	0.414764079147641\\
1.79333333333333	0.414193302891933\\
1.798	0.41324200913242\\
1.814	0.412861491628615\\
1.82466666666667	0.411149162861492\\
1.84866666666667	0.411149162861492\\
1.85066666666667	0.410578386605784\\
};
\addlegendentry{(59.2\%)};

\addplot [color=mycolor4,dotted,line width=2.0pt]
  table[row sep=crcr]{%
2.603	0.274923896499239\\
2.57933333333333	0.275684931506849\\
2.55766666666667	0.27796803652968\\
2.536	0.278919330289193\\
2.526	0.280251141552511\\
2.49466666666667	0.281773211567732\\
2.48633333333333	0.282914764079148\\
2.46766666666667	0.284056316590563\\
2.44533333333333	0.284627092846271\\
2.416	0.287290715372907\\
2.415	0.28824200913242\\
2.39366666666667	0.289193302891933\\
2.38233333333333	0.290715372907154\\
2.37033333333333	0.290905631659056\\
2.36366666666667	0.292047184170472\\
2.35333333333333	0.292237442922374\\
2.339	0.293759512937595\\
2.32466666666667	0.296613394216134\\
2.29	0.298896499238965\\
2.28	0.300989345509893\\
2.255	0.302891933028919\\
2.222	0.303843226788432\\
2.21566666666667	0.304794520547945\\
2.19033333333333	0.306126331811263\\
2.17166666666667	0.308599695585997\\
2.159	0.309741248097412\\
2.12133333333333	0.311643835616438\\
2.10733333333333	0.313165905631659\\
2.08666666666667	0.313736681887367\\
2.082	0.31468797564688\\
2.06	0.3162100456621\\
2.00466666666667	0.318873668188737\\
1.98566666666667	0.321156773211568\\
1.96966666666667	0.321537290715373\\
1.95766666666667	0.323630136986301\\
1.94866666666667	0.324010654490107\\
1.93833333333333	0.32572298325723\\
1.926	0.326674277016743\\
1.89866666666667	0.327625570776256\\
1.879	0.329528158295282\\
1.85533333333333	0.331050228310502\\
1.83466666666667	0.33162100456621\\
1.81866666666667	0.332572298325723\\
1.81233333333333	0.333713850837139\\
1.776	0.335806697108067\\
1.772	0.33675799086758\\
1.75033333333333	0.338089802130898\\
1.733	0.339992389649924\\
1.71766666666667	0.341133942161339\\
1.7	0.34341704718417\\
1.67833333333333	0.343987823439878\\
1.66766666666667	0.344939117199391\\
1.65233333333333	0.347222222222222\\
1.62533333333333	0.347983257229833\\
1.597	0.349695585996956\\
1.59333333333333	0.350646879756469\\
1.58033333333333	0.351217656012177\\
1.578	0.352168949771689\\
1.55766666666667	0.354452054794521\\
1.54633333333333	0.355213089802131\\
1.53166666666667	0.357496194824962\\
1.52466666666667	0.357686453576865\\
1.519	0.359208523592085\\
1.49833333333333	0.360730593607306\\
1.468	0.363394216133942\\
1.43766666666667	0.367770167427702\\
1.425	0.368721461187215\\
1.39633333333333	0.36986301369863\\
1.38666666666667	0.371385083713851\\
1.369	0.372526636225266\\
1.34566666666667	0.374619482496195\\
1.33066666666667	0.377092846270928\\
1.32266666666667	0.377473363774734\\
1.31466666666667	0.378995433789954\\
1.30233333333333	0.379756468797565\\
1.29266666666667	0.381468797564688\\
1.275	0.382800608828006\\
1.26633333333333	0.384512937595129\\
1.25133333333333	0.386415525114155\\
1.239	0.389269406392694\\
1.21	0.391552511415525\\
1.20633333333333	0.392313546423135\\
1.18166666666667	0.393264840182648\\
1.16966666666667	0.395167427701674\\
1.149	0.395738203957382\\
1.12066666666667	0.397831050228311\\
1.11633333333333	0.398972602739726\\
1.096	0.399543378995434\\
1.05366666666667	0.404680365296804\\
1.04	0.407724505327245\\
1.00333333333333	0.41248097412481\\
0.980666666666667	0.413622526636225\\
0.960333333333333	0.416286149162861\\
0.941666666666667	0.417998477929985\\
0.940666666666667	0.418949771689498\\
0.929666666666667	0.421042617960426\\
0.902666666666667	0.424657534246575\\
0.89	0.427891933028919\\
0.870666666666667	0.430745814307458\\
0.859	0.431316590563166\\
0.847666666666667	0.432648401826484\\
0.834	0.435692541856925\\
0.798333333333333	0.437785388127854\\
0.771333333333333	0.441019786910198\\
0.756	0.445015220700152\\
0.738666666666667	0.448059360730594\\
0.719	0.449961948249619\\
0.703333333333333	0.452054794520548\\
0.692	0.453006088280061\\
0.667666666666667	0.45738203957382\\
0.653666666666667	0.46099695585997\\
0.647	0.461567732115677\\
0.640333333333333	0.463660578386606\\
0.629	0.465753424657534\\
0.605	0.46765601217656\\
0.601	0.469939117199391\\
0.593333333333333	0.471080669710807\\
0.586333333333333	0.47355403348554\\
0.561333333333333	0.476407914764079\\
0.555	0.477739726027397\\
0.552333333333333	0.480022831050228\\
0.544333333333333	0.482686453576865\\
0.535333333333333	0.48382800608828\\
0.521333333333333	0.487442922374429\\
0.497666666666667	0.489916286149163\\
0.495333333333333	0.491628614916286\\
0.482333333333333	0.493531202435312\\
0.462	0.49923896499239\\
0.443	0.502663622526636\\
0.431	0.507229832572298\\
0.418666666666667	0.509132420091324\\
0.416666666666667	0.510464231354642\\
0.402666666666667	0.512747336377473\\
0.398666666666667	0.514459665144597\\
0.383666666666667	0.515981735159817\\
0.374666666666667	0.519406392694064\\
0.363666666666667	0.521499238964992\\
0.362	0.523021308980213\\
0.35	0.525875190258752\\
0.338333333333333	0.527587519025875\\
0.337333333333333	0.528919330289193\\
0.326666666666667	0.530631659056317\\
0.325	0.532153729071537\\
0.308666666666667	0.535578386605784\\
0.303	0.538812785388128\\
0.296666666666667	0.539573820395738\\
0.285333333333333	0.543569254185692\\
0.267	0.546993911719939\\
0.264	0.54851598173516\\
0.244666666666667	0.552891933028919\\
0.231666666666667	0.558219178082192\\
0.223666666666667	0.562595129375951\\
0.202	0.567732115677321\\
0.189666666666667	0.574581430745814\\
0.181333333333333	0.580669710806697\\
0.171333333333333	0.58599695585997\\
0.163666666666667	0.59189497716895\\
0.154666666666667	0.593607305936073\\
0.151333333333333	0.59703196347032\\
0.139	0.602739726027397\\
0.136333333333333	0.605783866057839\\
0.127333333333333	0.609208523592085\\
0.119	0.615106544901065\\
0.116333333333333	0.615677321156773\\
0.109	0.623477929984779\\
0.1	0.627853881278539\\
0.0996666666666667	0.62937595129376\\
0.0876666666666667	0.635654490106545\\
0.0873333333333333	0.637366818873668\\
0.0786666666666667	0.646689497716895\\
0.0743333333333333	0.650304414003044\\
0.073	0.655060882800609\\
0.069	0.656773211567732\\
0.0653333333333333	0.661339421613394\\
0.063	0.667427701674277\\
0.058	0.674086757990868\\
0.0526666666666667	0.67941400304414\\
0.05	0.686453576864536\\
0.0456666666666667	0.692351598173516\\
0.0433333333333333	0.699581430745814\\
0.041	0.702625570776256\\
0.039	0.711567732115677\\
0.0343333333333333	0.718797564687976\\
0.0333333333333333	0.72716894977169\\
0.0313333333333333	0.728310502283105\\
0.028	0.739916286149163\\
0.0266666666666667	0.741248097412481\\
0.0243333333333333	0.754756468797565\\
0.0226666666666667	0.757800608828006\\
0.021	0.770357686453577\\
0.016	0.78900304414003\\
0.0143333333333333	0.80365296803653\\
0.0126666666666667	0.81031202435312\\
0.00933333333333333	0.842085235920852\\
0.006	0.868721461187215\\
0.004	0.898401826484018\\
0.003	0.903538812785388\\
0.00166666666666667	0.94482496194825\\
0	0.964421613394216\\
0	0.999809741248097\\
};
\addlegendentry{(61.4\%)};

\end{axis}
\end{tikzpicture}%

%% file: 20170113b_sketchupResults.tex
%
\definecolor{mycolor1}{rgb}{0.85098,0.32549,0.09804}%
\definecolor{mycolor2}{rgb}{0.49412,0.18431,0.55686}%
\definecolor{mycolor3}{rgb}{0.00000,0.44706,0.74118}%
\begin{tikzpicture}
\tikzstyle{dashed}=[dash pattern=on 6pt off 2pt]
\tikzstyle{dotted}=[dash pattern=on \pgflinewidth off 2pt]

\begin{axis}[%
width=0.951\figurewidth,
height=0.822\figurewidth,
at={(0\figurewidth,0\figurewidth)},
scale only axis,
xmode=log,
xmin=0.001,
xmax=1,
xlabel={fppw},
xmajorgrids,
scaled ticks=false,
tick label style={/pgf/number format/fixed},
ymode=log,
ymin=0.1,
ymax=1,
ytick={0.1,0.2,0.5,1},
yticklabels={{0.1},{0.2},{0.5},{1}},
ylabel={miss rate},
ymajorgrids,
axis background/.style={fill=white},
axis x line*=bottom,
axis y line*=left,
legend style={at={(1.03,0.5)},anchor=west ,legend cell align=left,align=left,draw=white!15!black}
]
\addplot [color=mycolor1,solid,line width=2pt]
  table[row sep=crcr]{%
0	1\\
0.000365497076023392	0.989819004524887\\
0.000365497076023392	0.947209653092006\\
0.00219298245614035	0.906862745098039\\
0.00292397660818713	0.904977375565611\\
0.00292397660818713	0.882352941176471\\
0.0043859649122807	0.868024132730015\\
0.0043859649122807	0.82843137254902\\
0.00548245614035088	0.81184012066365\\
0.00804093567251462	0.802790346907994\\
0.0087719298245614	0.786953242835596\\
0.0120614035087719	0.762820512820513\\
0.0120614035087719	0.753016591251885\\
0.0160818713450292	0.743589743589744\\
0.016812865497076	0.729638009049774\\
0.0182748538011696	0.720965309200603\\
0.0182748538011696	0.704374057315234\\
0.0211988304093567	0.693438914027149\\
0.0244883040935673	0.671945701357466\\
0.0266812865497076	0.666289592760181\\
0.0266812865497076	0.657993966817496\\
0.0292397660818713	0.652337858220211\\
0.0339912280701754	0.613876319758673\\
0.0391081871345029	0.600678733031674\\
0.0391081871345029	0.591251885369532\\
0.0427631578947368	0.568627450980392\\
0.0471491228070175	0.556938159879336\\
0.0471491228070175	0.551659125188537\\
0.0515350877192982	0.542609351432881\\
0.0529970760233918	0.535444947209653\\
0.0559210526315789	0.533182503770739\\
0.0566520467836257	0.526772247360483\\
0.0595760233918129	0.524886877828054\\
0.0610380116959064	0.518476621417798\\
0.0646929824561404	0.515460030165912\\
0.0665204678362573	0.508295625942685\\
0.0716374269005848	0.5026395173454\\
0.0720029239766082	0.493589743589744\\
0.0730994152046784	0.493589743589744\\
0.0763888888888889	0.480769230769231\\
0.0818713450292398	0.47209653092006\\
0.0826023391812866	0.460407239819005\\
0.0873538011695906	0.454374057315234\\
0.0880847953216374	0.447586726998492\\
0.0950292397660819	0.438914027149321\\
0.0964912280701754	0.430618401206637\\
0.100511695906433	0.425716440422323\\
0.107090643274854	0.421191553544495\\
0.108918128654971	0.411764705882353\\
0.112207602339181	0.40422322775264\\
0.113669590643275	0.381975867269985\\
0.115131578947368	0.375565610859729\\
0.122076023391813	0.363876319758673\\
0.12609649122807	0.360859728506787\\
0.130482456140351	0.346907993966817\\
0.139619883040936	0.334087481146305\\
0.141812865497076	0.328054298642534\\
0.14875730994152	0.321266968325792\\
0.152046783625731	0.314102564102564\\
0.155701754385965	0.312971342383107\\
0.161184210526316	0.301282051282051\\
0.16483918128655	0.299773755656109\\
0.16812865497076	0.291478129713424\\
0.177997076023392	0.285444947209653\\
0.181286549707602	0.275641025641026\\
0.186038011695906	0.271870286576169\\
0.190789473684211	0.262066365007541\\
0.198099415204678	0.259049773755656\\
0.201388888888889	0.254524886877828\\
0.201388888888889	0.248868778280543\\
0.20577485380117	0.246606334841629\\
0.207602339181287	0.241704374057315\\
0.211622807017544	0.240950226244344\\
0.218932748538012	0.23340874811463\\
0.222222222222222	0.229638009049774\\
0.225511695906433	0.223981900452489\\
0.228801169590643	0.223981900452489\\
0.232821637426901	0.216817496229261\\
0.240497076023392	0.21606334841629\\
0.241228070175439	0.213046757164404\\
0.250365497076023	0.207390648567119\\
0.255482456140351	0.201734539969834\\
0.258771929824561	0.201357466063348\\
0.262061403508772	0.196078431372549\\
0.266081871345029	0.195324283559578\\
0.269005847953216	0.191176470588235\\
0.275584795321637	0.188536953242836\\
0.275584795321637	0.185897435897436\\
0.281067251461988	0.180618401206637\\
0.284722222222222	0.180241327300151\\
0.28874269005848	0.174585218702866\\
0.29422514619883	0.173831070889894\\
0.296052631578947	0.170060331825038\\
0.302266081871345	0.170060331825038\\
0.304824561403509	0.16553544494721\\
0.309941520467836	0.164027149321267\\
0.312865497076023	0.159879336349925\\
0.319809941520468	0.156485671191554\\
0.327850877192982	0.155354449472097\\
0.338084795321637	0.149321266968326\\
0.341739766081871	0.148567119155354\\
0.350146198830409	0.139517345399698\\
0.354897660818713	0.136123680241327\\
0.37280701754386	0.127828054298643\\
0.374269005847953	0.124811463046757\\
0.379385964912281	0.124057315233786\\
0.380847953216374	0.121417797888386\\
0.388523391812865	0.119532428355958\\
0.39875730994152	0.113499245852187\\
0.41812865497076	0.107088989441931\\
0.423245614035088	0.106711915535445\\
0.433479532163743	0.102941176470588\\
0.43640350877193	0.0991704374057315\\
0.441520467836257	0.0984162895927602\\
0.446637426900585	0.0927601809954751\\
0.455043859649123	0.0920060331825038\\
0.474049707602339	0.0871040723981901\\
0.475146198830409	0.0844645550527904\\
0.485745614035088	0.0844645550527904\\
0.491959064327485	0.0806938159879337\\
0.509502923976608	0.079185520361991\\
0.555921052631579	0.0641025641025641\\
0.570906432748538	0.0603318250377074\\
0.591008771929825	0.057315233785822\\
0.609283625730994	0.0565610859728507\\
0.612938596491228	0.053921568627451\\
0.629385964912281	0.05052790346908\\
0.635964912280702	0.0475113122171946\\
0.686769005847953	0.0377073906485671\\
0.6875	0.0358220211161387\\
0.701754385964912	0.0335595776772247\\
0.707236842105263	0.0301659125188537\\
0.725877192982456	0.0279034690799397\\
0.734283625730994	0.0256410256410257\\
0.767909356725146	0.02526395173454\\
0.77156432748538	0.0233785822021116\\
0.787646198830409	0.0211161387631976\\
0.795321637426901	0.0211161387631976\\
0.807017543859649	0.0173453996983409\\
0.816154970760234	0.0158371040723982\\
0.83077485380117	0.0154600301659125\\
0.838815789473684	0.0135746606334841\\
0.857456140350877	0.0131975867269984\\
0.881944444444444	0.00867269984917041\\
0.907529239766082	0.00867269984917041\\
0.921418128654971	0.00791855203619907\\
0.923245614035088	0.00641025641025639\\
0.93859649122807	0.00490196078431371\\
0.96125730994152	0.00452488687782804\\
0.967836257309941	0.00188536953242835\\
};
\addlegendentry{$0\rightarrow\sfrac{\pi}{4}$};

\addplot [color=mycolor2,solid,line width=2pt]
  table[row sep=crcr]{%
0	1\\
0	0.949095022624434\\
0.000730994152046784	0.945701357466063\\
0.000730994152046784	0.866138763197587\\
0.00182748538011696	0.855580693815988\\
0.00182748538011696	0.836349924585219\\
0.00219298245614035	0.836349924585219\\
0.00219298245614035	0.808446455505279\\
0.00292397660818713	0.793740573152338\\
0.00292397660818713	0.762443438914027\\
0.00328947368421053	0.762443438914027\\
0.00328947368421053	0.741704374057315\\
0.00402046783625731	0.737556561085973\\
0.00402046783625731	0.709653092006033\\
0.00548245614035088	0.70211161387632\\
0.00584795321637427	0.680995475113122\\
0.00657894736842105	0.678733031674208\\
0.00657894736842105	0.664404223227753\\
0.00730994152046784	0.662141779788839\\
0.00730994152046784	0.63499245852187\\
0.0087719298245614	0.633107088989442\\
0.0087719298245614	0.624434389140271\\
0.010233918128655	0.620663650075415\\
0.0105994152046784	0.591251885369532\\
0.0116959064327485	0.58974358974359\\
0.0116959064327485	0.577300150829563\\
0.0127923976608187	0.572398190045249\\
0.0131578947368421	0.558069381598793\\
0.0146198830409357	0.546757164404223\\
0.0146198830409357	0.539592760180996\\
0.0160818713450292	0.535067873303167\\
0.0182748538011696	0.515082956259427\\
0.0193713450292398	0.515082956259427\\
0.0211988304093567	0.498491704374057\\
0.0219298245614035	0.498491704374057\\
0.0219298245614035	0.478883861236802\\
0.0241228070175439	0.465686274509804\\
0.0263157894736842	0.462292609351433\\
0.028874269005848	0.448340874811463\\
0.028874269005848	0.437028657616893\\
0.0361842105263158	0.412141779788839\\
0.0394736842105263	0.396681749622926\\
0.0398391812865497	0.380844645550528\\
0.0434941520467836	0.377073906485671\\
0.0453216374269006	0.3710407239819\\
0.0489766081871345	0.351809954751131\\
0.0504385964912281	0.35105580693816\\
0.0526315789473684	0.335595776772247\\
0.0562865497076023	0.326923076923077\\
0.0581140350877193	0.315987933634992\\
0.0592105263157895	0.315610859728507\\
0.0603070175438596	0.303544494720965\\
0.0650584795321637	0.294117647058823\\
0.0665204678362573	0.284313725490196\\
0.0690789473684211	0.27790346907994\\
0.0738304093567251	0.271493212669683\\
0.0741959064327485	0.265460030165912\\
0.0771198830409357	0.26395173453997\\
0.0778508771929825	0.259049773755656\\
0.0833333333333333	0.248114630467572\\
0.0833333333333333	0.243212669683258\\
0.087719298245614	0.234162895927602\\
0.0899122807017544	0.23340874811463\\
0.0928362573099415	0.220588235294118\\
0.0957602339181287	0.217948717948718\\
0.0994152046783626	0.209276018099548\\
0.10233918128655	0.207390648567119\\
0.110380116959064	0.19683257918552\\
0.116228070175439	0.193438914027149\\
0.116228070175439	0.18815987933635\\
0.119883040935673	0.18552036199095\\
0.121710526315789	0.180618401206637\\
0.134502923976608	0.169306184012066\\
0.136330409356725	0.164781297134238\\
0.140350877192982	0.162141779788839\\
0.14327485380117	0.156485671191554\\
0.149122807017544	0.15422322775264\\
0.156432748538012	0.141779788838612\\
0.159722222222222	0.140648567119155\\
0.16483918128655	0.135369532428356\\
0.168494152046784	0.135369532428356\\
0.171783625730994	0.1289592760181\\
0.180190058479532	0.122926093514329\\
0.180190058479532	0.11500754147813\\
0.185672514619883	0.11236802413273\\
0.186769005847953	0.108220211161388\\
0.195175438596491	0.105203619909502\\
0.196271929824561	0.103318250377074\\
0.202485380116959	0.102187028657617\\
0.206505847953216	0.0946455505279035\\
0.210891812865497	0.0912518853695324\\
0.216008771929825	0.0904977375565611\\
0.218201754385965	0.0867269984917044\\
0.22733918128655	0.083710407239819\\
0.227704678362573	0.0810708898944194\\
0.233918128654971	0.0810708898944194\\
0.241593567251462	0.0757918552036199\\
0.253289473684211	0.0705128205128205\\
0.256944444444444	0.0652337858220211\\
0.26827485380117	0.0622171945701357\\
0.274122807017544	0.0580693815987934\\
0.281067251461988	0.0554298642533937\\
0.293859649122807	0.0535444947209653\\
0.300438596491228	0.0497737556561086\\
0.307383040935673	0.0486425339366516\\
0.307383040935673	0.0433634992458521\\
0.320175438596491	0.0392156862745098\\
0.332602339181287	0.0388386123680241\\
0.33296783625731	0.0377073906485671\\
0.343932748538012	0.0377073906485671\\
0.351973684210526	0.0365761689291101\\
0.357821637426901	0.0343137254901961\\
0.373538011695906	0.0335595776772247\\
0.382309941520468	0.0312971342383107\\
0.385964912280702	0.028657616892911\\
0.399488304093567	0.028657616892911\\
0.401681286549708	0.0256410256410257\\
0.418494152046784	0.023001508295626\\
0.424342105263158	0.0199849170437406\\
0.436769005847953	0.0180995475113123\\
0.448099415204678	0.0180995475113123\\
0.449195906432749	0.0165912518853696\\
0.460891812865497	0.0165912518853696\\
0.466374269005848	0.0154600301659125\\
0.48062865497076	0.0150829562594268\\
0.485014619883041	0.0139517345399698\\
0.499269005847953	0.0135746606334841\\
0.506213450292398	0.0124434389140271\\
0.543859649122807	0.0120663650075414\\
0.56140350877193	0.0109351432880844\\
0.562865497076023	0.00980392156862742\\
0.574195906432749	0.00942684766214175\\
0.590277777777778	0.00791855203619907\\
0.590277777777778	0.00716440422322773\\
0.623538011695906	0.00716440422322773\\
0.633771929824561	0.00565610859728505\\
0.636695906432749	0.00414781297134237\\
0.652777777777778	0.00339366515837103\\
0.691154970760234	0.00339366515837103\\
0.698830409356725	0.00263951734539969\\
0.738304093567251	0.00188536953242835\\
0.738304093567251	0.00150829562594268\\
0.792397660818713	0.00150829562594268\\
0.792397660818713	0.00113122171945701\\
0.819078947368421	0.00113122171945701\\
0.819078947368421	0.00075414781297134\\
0.837353801169591	0.00075414781297134\\
0.837353801169591	0.00037707390648567\\
0.880482456140351	0.00037707390648567\\
0.880482456140351	0\\
};
\addlegendentry{$0\rightarrow\sfrac{\pi}{4}$, im. warp.};

\addplot [color=mycolor3,solid,line width=2pt]
  table[row sep=crcr]{%
0	1\\
0	0.976998491704374\\
0.000365497076023392	0.976998491704374\\
0.000365497076023392	0.89630467571644\\
0.000730994152046784	0.89630467571644\\
0.000730994152046784	0.877450980392157\\
0.00146198830409357	0.872926093514329\\
0.00146198830409357	0.856711915535445\\
0.00219298245614035	0.848039215686274\\
0.00219298245614035	0.813725490196078\\
0.00292397660818713	0.809577677224736\\
0.00292397660818713	0.769230769230769\\
0.0043859649122807	0.751885369532428\\
0.0043859649122807	0.70211161387632\\
0.00548245614035088	0.688536953242836\\
0.00548245614035088	0.676470588235294\\
0.00804093567251462	0.659125188536953\\
0.00804093567251462	0.644796380090498\\
0.0120614035087719	0.604826546003017\\
0.0120614035087719	0.588989441930618\\
0.0160818713450292	0.576168929110106\\
0.016812865497076	0.558069381598793\\
0.0182748538011696	0.54788838612368\\
0.0182748538011696	0.525641025641026\\
0.0193713450292398	0.512820512820513\\
0.0233918128654971	0.498114630467572\\
0.025219298245614	0.480769230769231\\
0.0266812865497076	0.477752639517345\\
0.0266812865497076	0.466440422322775\\
0.0292397660818713	0.446078431372549\\
0.0307017543859649	0.440045248868778\\
0.0317982456140351	0.422699849170437\\
0.0339912280701754	0.412141779788839\\
0.0391081871345029	0.394796380090498\\
0.0391081871345029	0.382730015082956\\
0.0402046783625731	0.370663650075415\\
0.0420321637426901	0.367269984917044\\
0.043859649122807	0.352941176470588\\
0.0467836257309941	0.349547511312217\\
0.0471491228070175	0.340497737556561\\
0.0493421052631579	0.332202111613876\\
0.0515350877192982	0.329185520361991\\
0.0529970760233918	0.320135746606335\\
0.0559210526315789	0.318627450980392\\
0.0566520467836257	0.311085972850679\\
0.0595760233918129	0.308446455505279\\
0.0625	0.298642533936652\\
0.0646929824561404	0.297134238310709\\
0.0665204678362573	0.28921568627451\\
0.0683479532163743	0.28921568627451\\
0.0716374269005848	0.279034690799397\\
0.0720029239766082	0.271870286576169\\
0.0749269005847953	0.265082956259427\\
0.0760233918128655	0.257164404223228\\
0.0818713450292398	0.252262443438914\\
0.0826023391812866	0.244343891402715\\
0.0873538011695906	0.239064856711916\\
0.089546783625731	0.233031674208145\\
0.0910087719298246	0.233031674208145\\
0.0924707602339181	0.22737556561086\\
0.0953947368421053	0.225490196078431\\
0.0972222222222222	0.218325791855204\\
0.101608187134503	0.213800904977376\\
0.10672514619883	0.211538461538462\\
0.107821637426901	0.207390648567119\\
0.112938596491228	0.201357466063348\\
0.115131578947368	0.187782805429864\\
0.116959064327485	0.187405731523379\\
0.120248538011696	0.180618401206637\\
0.125	0.177601809954751\\
0.12719298245614	0.170060331825038\\
0.136330409356725	0.161387631975867\\
0.137061403508772	0.157239819004525\\
0.141081871345029	0.15422322775264\\
0.152046783625731	0.142156862745098\\
0.155701754385965	0.141779788838612\\
0.158625730994152	0.133861236802413\\
0.16483918128655	0.130090497737557\\
0.167763157894737	0.124811463046757\\
0.173245614035088	0.122171945701357\\
0.181286549707602	0.113876319758673\\
0.1875	0.110859728506787\\
0.191154970760234	0.106711915535445\\
0.198464912280702	0.105203619909502\\
0.203581871345029	0.0999245852187028\\
0.207236842105263	0.0991704374057315\\
0.211622807017544	0.0953996983408748\\
0.214912280701754	0.0950226244343891\\
0.222587719298246	0.0882352941176471\\
0.225511695906433	0.0874811463046757\\
0.225877192982456	0.0844645550527904\\
0.229897660818713	0.083710407239819\\
0.233187134502924	0.079185520361991\\
0.239400584795322	0.0780542986425339\\
0.242690058479532	0.0754147812971342\\
0.24890350877193	0.0750377073906485\\
0.258040935672515	0.0716440422322775\\
0.260233918128655	0.0682503770739065\\
0.267178362573099	0.0678733031674208\\
0.278508771929825	0.0656108597285068\\
0.280701754385965	0.0622171945701357\\
0.286915204678363	0.0599547511312217\\
0.293494152046784	0.0595776772247361\\
0.297149122807018	0.0558069381598794\\
0.308479532163743	0.0509049773755657\\
0.319809941520468	0.049396681749623\\
0.325292397660819	0.0456259426847662\\
0.346125730994152	0.0422322775263951\\
0.355263157894737	0.0373303167420814\\
0.365131578947368	0.0350678733031674\\
0.370248538011696	0.0301659125188537\\
0.37719298245614	0.029788838612368\\
0.378654970760234	0.0282805429864253\\
0.388523391812865	0.0279034690799397\\
0.388523391812865	0.0271493212669683\\
0.404239766081871	0.0260180995475113\\
0.405701754385965	0.0237556561085973\\
0.413377192982456	0.023001508295626\\
0.41374269005848	0.021870286576169\\
0.422149122807018	0.0196078431372549\\
0.430921052631579	0.0188536953242836\\
0.436769005847953	0.0158371040723982\\
0.445540935672515	0.0143288084464555\\
0.455409356725146	0.0143288084464555\\
0.461988304093567	0.0131975867269984\\
0.491959064327485	0.0120663650075414\\
0.500365497076023	0.0105580693815988\\
0.514619883040936	0.0105580693815988\\
0.519005847953216	0.00980392156862742\\
0.540204678362573	0.00980392156862742\\
0.54422514619883	0.00867269984917041\\
0.554824561403509	0.00867269984917041\\
0.569078947368421	0.0075414781297134\\
0.58516081871345	0.0075414781297134\\
0.58516081871345	0.00678733031674206\\
0.616228070175439	0.00678733031674206\\
0.624634502923977	0.00565610859728505\\
0.646929824561403	0.00527903469079938\\
0.64875730994152	0.00452488687782804\\
0.665570175438597	0.00452488687782804\\
0.672514619883041	0.0037707390648567\\
0.71016081871345	0.0037707390648567\\
0.71016081871345	0.00339366515837103\\
0.728070175438597	0.00301659125188536\\
0.728070175438597	0.00263951734539969\\
0.755847953216374	0.00226244343891402\\
0.764619883040936	0.00150829562594268\\
0.804093567251462	0.00150829562594268\\
0.804093567251462	0.00113122171945701\\
0.865862573099415	0.00113122171945701\\
};
\addlegendentry{$0\rightarrow\sfrac{\pi}{4}$, class. adapt.};

\addplot [color=mycolor1,dashed,line width=2pt]
  table[row sep=crcr]{%
0	1\\
0	0.874434389140271\\
0.000365497076023392	0.874434389140271\\
0.000365497076023392	0.848039215686274\\
0.000730994152046784	0.848039215686274\\
0.000730994152046784	0.807692307692308\\
0.00109649122807018	0.807692307692308\\
0.00109649122807018	0.765460030165912\\
0.00146198830409357	0.765460030165912\\
0.00146198830409357	0.742835595776772\\
0.00219298245614035	0.742081447963801\\
0.00328947368421053	0.714555052790347\\
0.00328947368421053	0.691930618401207\\
0.00475146198830409	0.676470588235294\\
0.00475146198830409	0.644796380090498\\
0.00511695906432749	0.644796380090498\\
0.00548245614035088	0.627073906485671\\
0.00584795321637427	0.627073906485671\\
0.00584795321637427	0.590497737556561\\
0.00621345029239766	0.575414781297134\\
0.00767543859649123	0.567496229260935\\
0.00767543859649123	0.539969834087481\\
0.0087719298245614	0.53657616892911\\
0.00913742690058479	0.502262443438914\\
0.0105994152046784	0.500377073906486\\
0.0120614035087719	0.479638009049774\\
0.0135233918128655	0.479260935143288\\
0.0142543859649123	0.454374057315234\\
0.0153508771929825	0.447586726998492\\
0.0160818713450292	0.421945701357466\\
0.016812865497076	0.415912518853695\\
0.016812865497076	0.399321266968326\\
0.0182748538011696	0.397812971342383\\
0.0182748538011696	0.389517345399698\\
0.0204678362573099	0.3789592760181\\
0.0208333333333333	0.361990950226244\\
0.0219298245614035	0.361236802413273\\
0.0222953216374269	0.341628959276018\\
0.0248538011695906	0.33710407239819\\
0.0259502923976608	0.329939668174962\\
0.027046783625731	0.329939668174962\\
0.0285087719298246	0.316742081447964\\
0.033625730994152	0.307315233785822\\
0.035453216374269	0.292232277526395\\
0.035453216374269	0.285444947209653\\
0.037280701754386	0.281674208144796\\
0.0380116959064327	0.268099547511312\\
0.0391081871345029	0.265460030165912\\
0.0405701754385965	0.249245852187029\\
0.0420321637426901	0.241704374057315\\
0.045687134502924	0.239064856711916\\
0.0508040935672515	0.228506787330317\\
0.0519005847953216	0.22209653092006\\
0.0548245614035088	0.220588235294118\\
0.0559210526315789	0.214932126696833\\
0.0573830409356725	0.214932126696833\\
0.0603070175438596	0.200226244343891\\
0.0614035087719298	0.200226244343891\\
0.0628654970760234	0.191930618401207\\
0.0690789473684211	0.180618401206637\\
0.0723684210526316	0.178355957767723\\
0.0730994152046784	0.168174962292609\\
0.0756578947368421	0.164781297134238\\
0.081140350877193	0.161764705882353\\
0.0866228070175439	0.150829562594268\\
0.091374269005848	0.145550527903469\\
0.100877192982456	0.141025641025641\\
0.101608187134503	0.137254901960784\\
0.104897660818713	0.133861236802413\\
0.109283625730994	0.133484162895928\\
0.112207602339181	0.127073906485671\\
0.114400584795322	0.126696832579186\\
0.114766081871345	0.119532428355958\\
0.118055555555556	0.118024132730015\\
0.119517543859649	0.113122171945701\\
0.125365497076023	0.108220211161388\\
0.129020467836257	0.103318250377074\\
0.137061403508772	0.0969079939668175\\
0.141081871345029	0.0901206636500754\\
0.145102339181287	0.0893665158371041\\
0.150950292397661	0.083710407239819\\
0.156432748538012	0.0822021116138764\\
0.157894736842105	0.0795625942684767\\
0.16812865497076	0.0750377073906485\\
0.168859649122807	0.0731523378582202\\
0.173976608187135	0.0727752639517345\\
0.173976608187135	0.0712669683257918\\
0.187865497076023	0.0671191553544495\\
0.187865497076023	0.0652337858220211\\
0.203581871345029	0.0595776772247361\\
0.213450292397661	0.0580693815987934\\
0.218201754385965	0.055052790346908\\
0.223684210526316	0.055052790346908\\
0.22624269005848	0.0531674208144797\\
0.232821637426901	0.0531674208144797\\
0.233187134502924	0.0520361990950227\\
0.239766081871345	0.051659125188537\\
0.244517543859649	0.0497737556561086\\
0.262792397660819	0.0475113122171946\\
0.263888888888889	0.0460030165912518\\
0.281067251461988	0.0433634992458521\\
0.282163742690059	0.0414781297134238\\
0.293494152046784	0.0407239819004525\\
0.302997076023392	0.0388386123680241\\
0.317251461988304	0.0384615384615384\\
0.326023391812865	0.0373303167420814\\
0.328947368421053	0.0354449472096531\\
0.357821637426901	0.0328054298642534\\
0.366593567251462	0.0301659125188537\\
0.374634502923977	0.0301659125188537\\
0.374634502923977	0.0294117647058824\\
0.389254385964912	0.028657616892911\\
0.394736842105263	0.0271493212669683\\
0.413011695906433	0.026395173453997\\
0.416666666666667	0.0245098039215687\\
0.446637426900585	0.0214932126696833\\
0.450657894736842	0.0199849170437406\\
0.479897660818713	0.0192307692307693\\
0.484649122807018	0.0177224736048266\\
0.510964912280702	0.0173453996983409\\
0.511695906432749	0.0165912518853696\\
0.538011695906433	0.0154600301659125\\
0.543859649122807	0.0143288084464555\\
0.558114035087719	0.0143288084464555\\
0.571271929824561	0.0128205128205128\\
0.588450292397661	0.0128205128205128\\
0.588450292397661	0.0124434389140271\\
0.60672514619883	0.0124434389140271\\
0.616959064327485	0.0116892911010558\\
0.631213450292398	0.0116892911010558\\
0.637061403508772	0.0101809954751131\\
0.659722222222222	0.00829562594268474\\
0.695175438596491	0.00829562594268474\\
0.697368421052632	0.00716440422322773\\
0.72624269005848	0.00716440422322773\\
0.726608187134503	0.00641025641025639\\
0.753289473684211	0.00641025641025639\\
0.759502923976608	0.00565610859728505\\
0.782529239766082	0.00527903469079938\\
0.787646198830409	0.00452488687782804\\
0.812865497076023	0.00452488687782804\\
0.820175438596491	0.00414781297134237\\
0.828947368421053	0.00226244343891402\\
0.841374269005848	0.00188536953242835\\
0.870614035087719	0.00188536953242835\\
0.881578947368421	0.00113122171945701\\
0.908260233918129	0.00113122171945701\\
0.911184210526316	0.00037707390648567\\
0.926900584795322	0.00037707390648567\\
0.926900584795322	0\\
};
\addlegendentry{$\sfrac{\pi}{8}\rightarrow\sfrac{\pi}{4}$};

\addplot [color=mycolor2,dashed,line width=2pt]
  table[row sep=crcr]{%
0	1\\
0	0.880090497737557\\
0.000365497076023392	0.880090497737557\\
0.000365497076023392	0.815610859728507\\
0.000730994152046784	0.815610859728507\\
0.000730994152046784	0.726244343891403\\
0.00109649122807018	0.726244343891403\\
0.00109649122807018	0.710030165912519\\
0.00146198830409357	0.710030165912519\\
0.00182748538011696	0.686274509803922\\
0.00219298245614035	0.686274509803922\\
0.00219298245614035	0.661387631975867\\
0.00292397660818713	0.651206636500754\\
0.00292397660818713	0.625942684766214\\
0.0043859649122807	0.610859728506787\\
0.0043859649122807	0.598039215686274\\
0.00475146198830409	0.598039215686274\\
0.00475146198830409	0.570889894419306\\
0.00511695906432749	0.570889894419306\\
0.00511695906432749	0.545248868778281\\
0.00548245614035088	0.545248868778281\\
0.00548245614035088	0.518853695324283\\
0.00584795321637427	0.518853695324283\\
0.00584795321637427	0.506787330316742\\
0.00657894736842105	0.506033182503771\\
0.00657894736842105	0.492081447963801\\
0.00730994152046784	0.491704374057315\\
0.00913742690058479	0.473227752639517\\
0.0105994152046784	0.46342383107089\\
0.0105994152046784	0.452488687782805\\
0.0116959064327485	0.45211161387632\\
0.0120614035087719	0.434389140271493\\
0.0127923976608187	0.434389140271493\\
0.0127923976608187	0.411387631975867\\
0.0138888888888889	0.408748114630468\\
0.0142543859649123	0.38763197586727\\
0.0157163742690058	0.385746606334842\\
0.0157163742690058	0.380090497737557\\
0.0182748538011696	0.368024132730015\\
0.0182748538011696	0.360859728506787\\
0.0197368421052632	0.346530920060332\\
0.0201023391812865	0.346530920060332\\
0.0204678362573099	0.323529411764706\\
0.0219298245614035	0.322021116138763\\
0.0222953216374269	0.313348416289593\\
0.0244883040935673	0.302413273001508\\
0.025219298245614	0.294871794871795\\
0.0255847953216374	0.279034690799397\\
0.0259502923976608	0.279034690799397\\
0.0259502923976608	0.265082956259427\\
0.0266812865497076	0.265082956259427\\
0.0274122807017544	0.257164404223228\\
0.0299707602339181	0.251885369532428\\
0.0307017543859649	0.245852187028658\\
0.033625730994152	0.237556561085973\\
0.033625730994152	0.230769230769231\\
0.035453216374269	0.223604826546003\\
0.0365497076023392	0.223227752639517\\
0.0369152046783626	0.21078431372549\\
0.0409356725146199	0.195701357466063\\
0.0442251461988304	0.19079939668175\\
0.0464181286549708	0.177224736048265\\
0.0500730994152047	0.174962292609351\\
0.0529970760233918	0.166289592760181\\
0.0537280701754386	0.156108597285068\\
0.0551900584795322	0.155354449472097\\
0.0566520467836257	0.147058823529412\\
0.0614035087719298	0.141025641025641\\
0.0628654970760234	0.136877828054299\\
0.0665204678362573	0.133861236802413\\
0.0679824561403509	0.124057315233786\\
0.0738304093567251	0.116515837104072\\
0.0749269005847953	0.10972850678733\\
0.0771198830409357	0.106334841628959\\
0.0847953216374269	0.102187028657617\\
0.0873538011695906	0.0923831070889894\\
0.0932017543859649	0.0908748114630468\\
0.0968567251461988	0.084841628959276\\
0.103070175438596	0.0806938159879337\\
0.105263157894737	0.0754147812971342\\
0.107821637426901	0.0727752639517345\\
0.111111111111111	0.0720211161387632\\
0.113304093567251	0.0693815987933635\\
0.116593567251462	0.0690045248868778\\
0.120248538011696	0.0659879336349924\\
0.120248538011696	0.0633484162895928\\
0.123172514619883	0.0610859728506787\\
0.127558479532164	0.0607088989441931\\
0.133406432748538	0.0535444947209653\\
0.139254385964912	0.052790346907994\\
0.145102339181287	0.0497737556561086\\
0.146198830409357	0.0463800904977375\\
0.150950292397661	0.0441176470588235\\
0.155701754385965	0.0437405731523378\\
0.156798245614035	0.0411010558069381\\
0.165204678362573	0.0411010558069381\\
0.170687134502924	0.0395927601809954\\
0.170687134502924	0.0343137254901961\\
0.176900584795322	0.0328054298642534\\
0.184941520467836	0.0290346907993967\\
0.194078947368421	0.0290346907993967\\
0.200292397660819	0.026395173453997\\
0.208333333333333	0.0256410256410257\\
0.212719298245614	0.0233785822021116\\
0.222953216374269	0.0226244343891403\\
0.223318713450292	0.0214932126696833\\
0.231359649122807	0.0214932126696833\\
0.240131578947368	0.0199849170437406\\
0.246345029239766	0.0199849170437406\\
0.252558479532164	0.0188536953242836\\
0.259868421052632	0.0188536953242836\\
0.266081871345029	0.0173453996983409\\
0.275219298245614	0.0173453996983409\\
0.278874269005848	0.0158371040723982\\
0.295321637426901	0.0150829562594268\\
0.298611111111111	0.0135746606334841\\
0.310672514619883	0.0135746606334841\\
0.322733918128655	0.0128205128205128\\
0.323830409356725	0.0120663650075414\\
0.333333333333333	0.0120663650075414\\
0.338084795321637	0.0109351432880844\\
0.370979532163743	0.0109351432880844\\
0.385964912280702	0.00942684766214175\\
0.387792397660819	0.00867269984917041\\
0.41593567251462	0.00791855203619907\\
0.419956140350877	0.00716440422322773\\
0.469298245614035	0.00716440422322773\\
0.469298245614035	0.00678733031674206\\
0.482090643274854	0.00678733031674206\\
0.500730994152047	0.00565610859728505\\
0.555555555555556	0.00565610859728505\\
0.566154970760234	0.00490196078431371\\
0.584795321637427	0.00490196078431371\\
0.584795321637427	0.00452488687782804\\
0.596491228070175	0.00452488687782804\\
0.600877192982456	0.0037707390648567\\
0.641081871345029	0.0037707390648567\\
0.652412280701754	0.00301659125188536\\
0.685672514619883	0.00301659125188536\\
0.685672514619883	0.00263951734539969\\
0.700292397660819	0.00263951734539969\\
0.700292397660819	0.00226244343891402\\
0.83077485380117	0.00226244343891402\\
0.83077485380117	0.00188536953242835\\
0.863669590643275	0.00188536953242835\\
0.870979532163743	0.00075414781297134\\
0.886695906432749	0.00075414781297134\\
0.886695906432749	0.00037707390648567\\
0.904605263157895	0.00037707390648567\\
0.904605263157895	0\\
};
\addlegendentry{$\sfrac{\pi}{8}\rightarrow\sfrac{\pi}{4}$, im. warp.};

\addplot [color=mycolor3,dashed,line width=2pt]
  table[row sep=crcr]{%
0	1\\
0	0.78393665158371\\
0.000365497076023392	0.78393665158371\\
0.000365497076023392	0.737556561085973\\
0.000730994152046784	0.737556561085973\\
0.000730994152046784	0.691553544494721\\
0.00109649122807018	0.691553544494721\\
0.00109649122807018	0.639894419306184\\
0.00146198830409357	0.639894419306184\\
0.00146198830409357	0.608974358974359\\
0.00255847953216374	0.601432880844646\\
0.00292397660818713	0.576168929110106\\
0.00328947368421053	0.576168929110106\\
0.00328947368421053	0.556938159879336\\
0.00402046783625731	0.550904977375566\\
0.00402046783625731	0.540346907993967\\
0.00475146198830409	0.535822021116139\\
0.00475146198830409	0.502262443438914\\
0.00511695906432749	0.502262443438914\\
0.00511695906432749	0.488310708898944\\
0.00548245614035088	0.488310708898944\\
0.00548245614035088	0.474358974358974\\
0.00584795321637427	0.474358974358974\\
0.00584795321637427	0.435143288084465\\
0.00621345029239766	0.435143288084465\\
0.00621345029239766	0.413650075414781\\
0.00767543859649123	0.403469079939668\\
0.00767543859649123	0.383484162895928\\
0.0087719298245614	0.377073906485671\\
0.00913742690058479	0.34841628959276\\
0.0105994152046784	0.345022624434389\\
0.0105994152046784	0.339366515837104\\
0.0120614035087719	0.331825037707391\\
0.0120614035087719	0.326923076923077\\
0.0135233918128655	0.32579185520362\\
0.0142543859649123	0.303921568627451\\
0.0153508771929825	0.29524886877828\\
0.0160818713450292	0.277149321266968\\
0.016812865497076	0.274132730015083\\
0.016812865497076	0.259049773755656\\
0.0182748538011696	0.256787330316742\\
0.0182748538011696	0.245475113122172\\
0.0204678362573099	0.238310708898944\\
0.0208333333333333	0.226998491704374\\
0.0219298245614035	0.226998491704374\\
0.0222953216374269	0.21342383107089\\
0.0274122807017544	0.202488687782805\\
0.0296052631578947	0.194570135746606\\
0.0321637426900585	0.192684766214178\\
0.0347222222222222	0.185143288084465\\
0.035453216374269	0.174962292609351\\
0.0365497076023392	0.17420814479638\\
0.0380116959064327	0.162141779788839\\
0.0402046783625731	0.157239819004525\\
0.0405701754385965	0.147435897435897\\
0.0420321637426901	0.144042232277526\\
0.0449561403508772	0.141779788838612\\
0.0467836257309941	0.138009049773756\\
0.0508040935672515	0.133484162895928\\
0.0529970760233918	0.124434389140271\\
0.0551900584795322	0.123303167420814\\
0.0559210526315789	0.119909502262443\\
0.0581140350877193	0.119155354449472\\
0.0603070175438596	0.111990950226244\\
0.0614035087719298	0.111990950226244\\
0.0650584795321637	0.0995475113122172\\
0.0723684210526316	0.0961538461538461\\
0.0730994152046784	0.0904977375565611\\
0.0778508771929825	0.0878582202111614\\
0.0796783625730994	0.0844645550527904\\
0.0822368421052632	0.0840874811463047\\
0.0862573099415205	0.0799396681749623\\
0.0899122807017544	0.0795625942684767\\
0.091374269005848	0.0761689291101055\\
0.0964912280701754	0.0739064856711915\\
0.0964912280701754	0.0723981900452488\\
0.101608187134503	0.0712669683257918\\
0.10233918128655	0.0690045248868778\\
0.10672514619883	0.0667420814479638\\
0.10672514619883	0.0652337858220211\\
0.114400584795322	0.0614630467571644\\
0.114400584795322	0.0565610859728507\\
0.121710526315789	0.0512820512820513\\
0.127558479532164	0.0509049773755657\\
0.131578947368421	0.0490196078431373\\
0.132675438596491	0.0467571644042232\\
0.140716374269006	0.0429864253393665\\
0.141081871345029	0.0411010558069381\\
0.145102339181287	0.0407239819004525\\
0.150584795321637	0.0369532428355958\\
0.156432748538012	0.0361990950226244\\
0.157894736842105	0.0346907993966817\\
0.172514619883041	0.0343137254901961\\
0.182017543859649	0.0335595776772247\\
0.186769005847953	0.0324283559577677\\
0.193713450292398	0.0324283559577677\\
0.199926900584795	0.0305429864253394\\
0.202485380116959	0.028657616892911\\
0.224780701754386	0.0271493212669683\\
0.22733918128655	0.0256410256410257\\
0.240497076023392	0.0248868778280543\\
0.250730994152047	0.0233785822021116\\
0.260599415204678	0.0233785822021116\\
0.263888888888889	0.0222473604826546\\
0.281067251461988	0.0211161387631976\\
0.285453216374269	0.0199849170437406\\
0.298611111111111	0.0199849170437406\\
0.302266081871345	0.0188536953242836\\
0.31030701754386	0.0188536953242836\\
0.312134502923977	0.0177224736048266\\
0.327119883040936	0.0177224736048266\\
0.328216374269006	0.0165912518853696\\
0.341374269005848	0.0162141779788839\\
0.346125730994152	0.0150829562594268\\
0.357456140350877	0.0150829562594268\\
0.371710526315789	0.0131975867269984\\
0.374634502923977	0.0113122171945701\\
0.39327485380117	0.0113122171945701\\
0.395833333333333	0.0105580693815988\\
0.413377192982456	0.00942684766214175\\
0.415204678362573	0.00829562594268474\\
0.429093567251462	0.00829562594268474\\
0.4375	0.0075414781297134\\
0.451388888888889	0.0075414781297134\\
0.451388888888889	0.00716440422322773\\
0.493786549707602	0.00716440422322773\\
0.496345029239766	0.00641025641025639\\
0.521198830409357	0.00603318250377072\\
0.521198830409357	0.00565610859728505\\
0.54422514619883	0.00565610859728505\\
0.555921052631579	0.00490196078431371\\
0.555921052631579	0.00452488687782804\\
0.572002923976608	0.00452488687782804\\
0.577485380116959	0.0037707390648567\\
0.628654970760234	0.0037707390648567\\
0.628654970760234	0.00339366515837103\\
0.678362573099415	0.00339366515837103\\
0.678362573099415	0.00301659125188536\\
0.697368421052632	0.00301659125188536\\
0.705409356725146	0.00226244343891402\\
0.729897660818713	0.00226244343891402\\
0.729897660818713	0.00188536953242835\\
0.799342105263158	0.00188536953242835\\
0.799342105263158	0.00150829562594268\\
0.841739766081871	0.00150829562594268\\
0.842836257309941	0.00075414781297134\\
0.884868421052632	0.00075414781297134\\
0.884868421052632	0.00037707390648567\\
0.919590643274854	0.00037707390648567\\
0.919590643274854	0\\
};
\addlegendentry{$\sfrac{\pi}{8}\rightarrow\sfrac{\pi}{4}$, class. adapt.};

\addplot [color=mycolor1,dotted,line width=2pt]
  table[row sep=crcr]{%
0	1\\
0	0.737933634992459\\
0.000365497076023392	0.737933634992459\\
0.000365497076023392	0.576168929110106\\
0.000730994152046784	0.576168929110106\\
0.000730994152046784	0.558446455505279\\
0.00109649122807018	0.558446455505279\\
0.00109649122807018	0.544117647058824\\
0.00146198830409357	0.544117647058824\\
0.00146198830409357	0.535444947209653\\
0.00182748538011696	0.535444947209653\\
0.00182748538011696	0.518099547511312\\
0.00255847953216374	0.516591251885369\\
0.00255847953216374	0.47737556561086\\
0.00292397660818713	0.47737556561086\\
0.00292397660818713	0.468325791855204\\
0.00365497076023392	0.464932126696833\\
0.00365497076023392	0.439291101055807\\
0.00548245614035088	0.427224736048265\\
0.00548245614035088	0.375942684766214\\
0.00584795321637427	0.375942684766214\\
0.00584795321637427	0.366515837104072\\
0.00657894736842105	0.358597285067873\\
0.00767543859649123	0.353318250377074\\
0.00767543859649123	0.345399698340875\\
0.00840643274853801	0.338612368024133\\
0.00913742690058479	0.338235294117647\\
0.00950292397660819	0.332579185520362\\
0.00950292397660819	0.317496229260935\\
0.010233918128655	0.314856711915535\\
0.0105994152046784	0.303544494720965\\
0.0116959064327485	0.301659125188537\\
0.0116959064327485	0.288084464555053\\
0.0124269005847953	0.286199095022624\\
0.0124269005847953	0.266214177978884\\
0.0131578947368421	0.262066365007541\\
0.0131578947368421	0.2526395173454\\
0.0138888888888889	0.25\\
0.0138888888888889	0.239441930618401\\
0.0146198830409357	0.233031674208145\\
0.0157163742690058	0.232277526395173\\
0.0175438596491228	0.22473604826546\\
0.0179093567251462	0.216817496229261\\
0.0193713450292398	0.214932126696833\\
0.0201023391812865	0.205882352941177\\
0.0208333333333333	0.205882352941177\\
0.0208333333333333	0.200980392156863\\
0.0230263157894737	0.199849170437406\\
0.0241228070175439	0.197209653092006\\
0.0244883040935673	0.189291101055807\\
0.025219298245614	0.189291101055807\\
0.0255847953216374	0.180995475113122\\
0.0263157894736842	0.180995475113122\\
0.027046783625731	0.170814479638009\\
0.0281432748538012	0.170814479638009\\
0.0281432748538012	0.16289592760181\\
0.0296052631578947	0.153469079939668\\
0.0307017543859649	0.151206636500754\\
0.0310672514619883	0.142156862745098\\
0.0325292397660819	0.137254901960784\\
0.0343567251461988	0.134615384615385\\
0.035453216374269	0.129336349924585\\
0.0391081871345029	0.124057315233786\\
0.0391081871345029	0.117269984917044\\
0.0413011695906433	0.115761689291101\\
0.0449561403508772	0.108974358974359\\
0.0464181286549708	0.108974358974359\\
0.0478801169590643	0.102564102564103\\
0.0500730994152047	0.101809954751131\\
0.0504385964912281	0.0999245852187028\\
0.0529970760233918	0.0980392156862745\\
0.0551900584795322	0.0908748114630468\\
0.0588450292397661	0.0886123680241327\\
0.060672514619883	0.0844645550527904\\
0.0621345029239766	0.0844645550527904\\
0.0635964912280702	0.081447963800905\\
0.0661549707602339	0.0810708898944194\\
0.0672514619883041	0.0769230769230769\\
0.0694444444444444	0.0735294117647058\\
0.0767543859649123	0.0716440422322775\\
0.0778508771929825	0.0671191553544495\\
0.0818713450292398	0.0622171945701357\\
0.0873538011695906	0.0607088989441931\\
0.0899122807017544	0.0576923076923077\\
0.0928362573099415	0.0576923076923077\\
0.0939327485380117	0.0558069381598794\\
0.0990497076023392	0.0531674208144797\\
0.100146198830409	0.0512820512820513\\
0.103070175438596	0.0501508295625943\\
0.10453216374269	0.047134238310709\\
0.10672514619883	0.047134238310709\\
0.108187134502924	0.0448717948717948\\
0.112938596491228	0.0437405731523378\\
0.114400584795322	0.0414781297134238\\
0.121345029239766	0.0407239819004525\\
0.121345029239766	0.0395927601809954\\
0.125365497076023	0.0373303167420814\\
0.131578947368421	0.0365761689291101\\
0.133406432748538	0.0346907993966817\\
0.140350877192982	0.0346907993966817\\
0.152046783625731	0.0331825037707391\\
0.161184210526316	0.029788838612368\\
0.168494152046784	0.0294117647058824\\
0.171783625730994	0.027526395173454\\
0.190423976608187	0.02526395173454\\
0.195540935672515	0.0233785822021116\\
0.202119883040936	0.0233785822021116\\
0.202485380116959	0.0222473604826546\\
0.226608187134503	0.0199849170437406\\
0.226973684210526	0.0192307692307693\\
0.238304093567251	0.0192307692307693\\
0.254751461988304	0.0158371040723982\\
0.267543859649123	0.0150829562594268\\
0.271929824561404	0.0131975867269984\\
0.285818713450292	0.0128205128205128\\
0.288377192982456	0.0120663650075414\\
0.304824561403509	0.0113122171945701\\
0.308114035087719	0.0105580693815988\\
0.319078947368421	0.0105580693815988\\
0.323099415204678	0.00942684766214175\\
0.331505847953216	0.00942684766214175\\
0.336988304093567	0.00716440422322773\\
0.346856725146199	0.00716440422322773\\
0.346856725146199	0.00678733031674206\\
0.356359649122807	0.00641025641025639\\
0.356359649122807	0.00603318250377072\\
0.37280701754386	0.00565610859728505\\
0.37280701754386	0.00527903469079938\\
0.392543859649123	0.00527903469079938\\
0.392543859649123	0.00490196078431371\\
0.406067251461988	0.00490196078431371\\
0.406067251461988	0.00452488687782804\\
0.421783625730994	0.00452488687782804\\
0.426535087719298	0.0037707390648567\\
0.436038011695906	0.0037707390648567\\
0.437865497076023	0.00301659125188536\\
0.467836257309941	0.00301659125188536\\
0.468567251461988	0.00226244343891402\\
0.492324561403509	0.00226244343891402\\
0.492324561403509	0.00188536953242835\\
0.591739766081871	0.00188536953242835\\
0.591739766081871	0.00150829562594268\\
0.619152046783626	0.00150829562594268\\
0.619152046783626	0.00113122171945701\\
0.659722222222222	0.00113122171945701\\
0.659722222222222	0.00075414781297134\\
0.905336257309941	0.00075414781297134\\
0.905336257309941	0.00037707390648567\\
0.986111111111111	0.00037707390648567\\
0.986111111111111	0\\
};
\addlegendentry{$\sfrac{\pi}{4}\rightarrow\sfrac{\pi}{4}$};

\end{axis}
\end{tikzpicture}%

%% file: 20160308_sketchupRobustness.tex
%
\begin{tikzpicture}

\begin{axis}[%
width=\figurewidth,
height=0.528\figurewidth,
at={(0\figurewidth,0\figurewidth)},
scale only axis,
xmin=0,
xmax=1.6,
xlabel={estimated target elevation},
scaled ticks=false,
tick label style={/pgf/number format/fixed},
ymin=0.65,
ymax=1,
ylabel={miss rate @ fppw=$10^{-2}$},
axis background/.style={fill=white},
legend style={at={(0.03,0.97)},anchor=north west,legend cell align=left,align=left,draw=white!15!black}
]
\addplot [color=black,solid,line width=2.5pt,forget plot]
  table[row sep=crcr]{%
0.0436332312998583	0.786324786324786\\
0.0872664625997165	0.773881347410759\\
0.130899693899575	0.767596782302665\\
0.174532925199433	0.762317747611865\\
0.218166156499291	0.751382604323781\\
0.26179938779915	0.741578682755153\\
0.305432619099008	0.738436400201106\\
0.349065850398866	0.729763700351936\\
0.392699081698724	0.707264957264957\\
0.436332312998582	0.699220713926596\\
0.479965544298441	0.69683257918552\\
0.523598775598299	0.693187531422826\\
0.567232006898157	0.67948717948718\\
0.610865238198015	0.671694318753142\\
0.654498469497873	0.662393162393162\\
0.698131700797732	0.664278531925591\\
0.74176493209759	0.667672197083962\\
0.785398163397448	0.672322775263952\\
0.829031394697306	0.676470588235294\\
0.872664625997165	0.69947209653092\\
0.916297857297023	0.718954248366013\\
0.959931088596881	0.743966817496229\\
1.00356431989674	0.765208647561589\\
1.0471975511966	0.786450477626948\\
1.09083078249646	0.80718954248366\\
1.13446401379631	0.832704876822524\\
1.17809724509617	0.854952237305178\\
1.22173047639603	0.894167923579688\\
1.26536370769589	0.922448466566114\\
1.30899693899575	0.948466566113625\\
1.35263017029561	0.972473604826546\\
1.39626340159546	0.991075917546506\\
1.43989663289532	0.996229260935143\\
1.48352986419518	0.999371543489191\\
1.52716309549504	0.998366013071895\\
};
\addplot [color=black,dashed,line width=1.5pt]
  table[row sep=crcr]{%
0.0436332312998583	0.796380090497738\\
1.52716309549504	0.796380090497738\\
};
\addlegendentry{performance of unadaptated detector};

\addplot [color=black,dotted,line width=1.5pt]
  table[row sep=crcr]{%
0.785398163397448	0.672322775263952\\
0.785398163397448	0.999371543489191\\
};
\addlegendentry{true target elevation};

\end{axis}
\end{tikzpicture}%